\documentclass[lettersize,journal]{IEEEtran}
\usepackage{amsmath,amsfonts}
\usepackage{algorithmic}
\usepackage{
algorithm}
\usepackage{array}
\usepackage[caption=false,font=normalsize,labelfont=sf,textfont=sf]{subfig}
\usepackage{textcomp}
\usepackage{stfloats}
\usepackage{url}
\usepackage{verbatim}
\usepackage{graphicx}
\usepackage{cite}

\usepackage{threeparttable}
\usepackage{multirow}
\usepackage{booktabs}
\usepackage{makecell}
\usepackage{pifont}
\usepackage{url}
\usepackage{xcolor}
\usepackage{hyperref}
\hypersetup{hypertex=true,
	colorlinks=true,
	linkcolor=blue,
	anchorcolor=blue,
	citecolor=blue}
\usepackage{color}

\newcommand{\revise}[1]{\textcolor{black}{#1}}

\begin{document}

\title{Cross-Modal Learning for Anomaly Detection in Complex Industrial Process: Methodology and Benchmark}

\author{Gaochang Wu, Yapeng Zhang, Lan Deng, Jingxin Zhang,~\IEEEmembership{Member,~IEEE} and Tianyou Chai,~\IEEEmembership{Life Fellow,~IEEE}
\thanks{This work was supported in part by the National Natural Science Foundation of China (NSFC) under Grant 61991404, 62103092, 61991401, U20A20189, and 62173120, in part by the Research Program of the Liaoning Liaohe Laboratory, No. LLL23ZZ-05-01, and in part by the Fundamental Research Funds for the Central Universities, No. N2424004. \textit{(Corresponding Author: Tianyou Chai.)}}
\thanks{Gaochang Wu, Yapeng Zhang, Lan Deng and Tianyou Chai are with the State Key Laboratory of Synthetical Automation for Process Industries, Northeastern University, Shenyang 110819, P. R. China. E-mail: \{wugc, tychai\}@mail.neu.edu.cn.}
\thanks{Jingxin Zhang is with the School of Software and Electrical Engineering, Swinburne University of Technology, Melbourne, VIC 3122, Australia. E-mail: jingxinzhang@swin.edu.au.}

}

\markboth{Journal of \LaTeX\ Class Files,~Vol.~14, No.~8, August~2021}%
{Shell \MakeLowercase{\textit{et al.}}: A Sample Article Using IEEEtran.cls for IEEE Journals}


\maketitle

\begin{abstract}
\revise{Anomaly detection in complex industrial processes plays a pivotal role in ensuring efficient, stable, and secure operation.}
Existing anomaly detection methods primarily focus on analyzing dominant anomalies using the process variables (such as arc current) or constructing neural networks based on abnormal visual features, while overlooking the intrinsic correlation of cross-modal information. 
This paper proposes a cross-modal Transformer (dubbed FmFormer), designed to facilitate anomaly detection by exploring the correlation between visual features (video) and process variables (current) \revise{in the context of the fused magnesium smelting process}.
Our approach introduces a novel tokenization paradigm to effectively bridge the substantial dimensionality gap between the 3D video modality and the 1D current modality in a multiscale manner, enabling a hierarchical reconstruction of pixel-level anomaly detection.
Subsequently, the FmFormer leverages self-attention to learn internal features within each modality and bidirectional cross-attention to capture correlations across modalities.
By decoding the bidirectional correlation features, we obtain the final detection result and even locate the specific anomaly region.
To validate the effectiveness of the proposed method, we also present a pioneering cross-modal benchmark of the fused magnesium smelting process, featuring synchronously acquired video and current data for over 2.2 million samples. 
Leveraging cross-modal learning, the proposed FmFormer achieves state-of-the-art performance in detecting anomalies, particularly under extreme interferences such as current fluctuations and visual occlusion caused by heavy water mist.
The presented methodology and benchmark may be applicable to other industrial applications with some amendments.
The benchmark will be released at \url{https://github.com/GaochangWu/FMF-Benchmark}.
\end{abstract}


\section{Introduction}
\IEEEPARstart{F}{used} \revise{magnesium smelting is a typical complex industrial process that produces fused magnesia, which has numerous properties, including high temperature resistance, strong oxidation resistance, and corrosion resistance. These qualities make it an indispensable resource in fields such as clinical surgery, aerospace, and various industries~\cite{Wu2018NonlinearCT}.} Fused Magnesium Furnace (FMF) is the main equipment in the fused magnesium smelting process to melt powdered ore by generating a molten pool around 2850$^\circ$C through electric arc. Since the entire production is a continuous alternation of ore feeding and smelting, coupled with dynamic fluctuations in ore properties and non-optimal settings of smelting current~\cite{Chai2017ACB}, the temperature of molten pool can become unstable, leading to abnormal conditions in the FMF. Typical anomalies, especially the semi-molten condition~\cite{Wu2015DataDrivenAC}, are detrimental to product quality and even threaten production security if not resolved in time. Therefore, timely and accurate anomaly detection is essential to the high quality, stable, and secure production of the fused magnesium smelting process.


In consideration of the intrinsic formation mechanism of anomaly, early studies focus on the detection, diagnosis or identification of the smelting current anomaly using knowledge-based or data-driven approaches. For instance, Wu \textit{et al.}~\cite{Wu2015DataDrivenAC} introduced an abnormal condition identification method by constructing expert rules based on the features of current tracking error, current change rate, arc resistance, etc. Zhang \textit{et al.}~\cite{Zhang2011FaultDO} combined multiscale kernel principal component analysis and multiscale kernel partial least analysis to extract the dominant anomaly information in a smelting process. Wang \textit{et al.}~\cite{Wang2023DensityBasedSP} proposed a abnormal variable isolation method by projecting the main variables including three-phase current and voltage into a structure preserving space. 
However, the high-frequency fluctuation of the smelting current, stemming from variations in resistance and the unstable arcing distance induced by liquid tumbling, makes the accurate and reliable detection exceedingly challenging.

\begin{figure*}
	\begin{center}
		\includegraphics[width=0.99\linewidth]{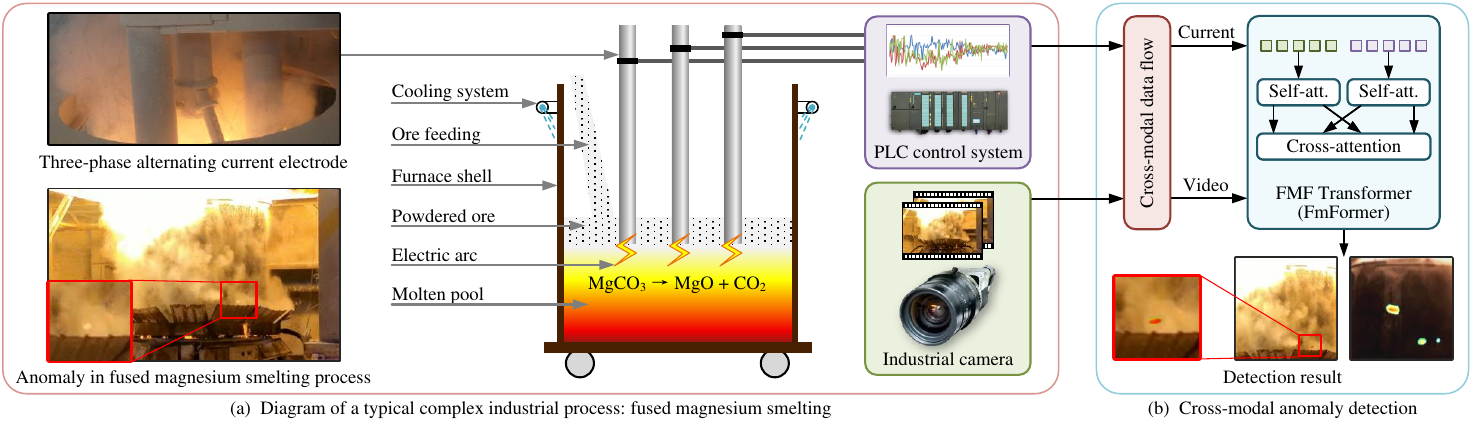}
	\end{center}
	\vspace{-3mm}
	\caption{Cross-modal information is exploited to \revise{perform anomaly detection in the context of a typical industrial process}, fused magnesium smelting, as illustrated in (a). The picture at the bottom left shows an anomaly region on the furnace shell, whose visual feature is difficult to detect due to interference from heavy water mist. A novel FMF Transformer (FmFormer) is proposed using synchronous acquired video and current data, to explore the internal features of each modality by self-attention and the correlation feature across modalities by cross-attention, as shown in (b).}
	\label{fig:FMF}
\vspace{-3mm}
\end{figure*}

With the success of deep learning in artificial intelligence~\cite{LeCun2015DeepL,Silver2017MasteringTG}, recent researches are stepping towards deep learning-based anomaly detection using image input~\cite{Zhou2022IdentificationOA,Lu2022SemiSupervisedCM,10587314} or video~\cite{wu2019abnormal,Liu2023DisturbanceRA} acquired from industrial cameras. Comparing with smelting current, abnormal conditions of FMF demonstrate prominent visual features. Taking the semi-molten condition as an example, in its initial phase, undesired variation in the molten pool temperature is accompanied by a red dot in the local area of the furnace shell~\cite{wu2019abnormal}. Based on this visual feature, Wu \textit{et al.}~\cite{wu2019abnormal} separated the anomaly detection task into the spatial feature extraction from 3D video using a 2D Convolutional Neural Network (CNN), and the temporal feature extraction and prediction using a Recurrent Neural Network (RNN).
Recently, Liu \textit{et al.}~\cite{Liu2023DisturbanceRA} further explored a 3D convolutional Long-Short Term Memory (LSTM)~\cite{Hochreiter1997LongSM} to learn spatio-temporal video data simultaneously. Although the vision-based anomaly detection provides a scheme with higher accuracy and better prediction consistency than the current-based detection, there are still challenges in such complex industrial environment due to visual interference. For instance, the detection accuracy can quickly degenerate when
there is water mist or light impact from the furnace flame (as shown in the bottom left image of Fig.~\ref{fig:FMF}(a)), which occasionally occur in such complex industrial scenario. Given these limitations of single-modality information, it is crucial to utilize both video and current information for a comprehensive anomaly detection in fused magnesium smelting processes.

To enhance anomaly detection in the context of the fused magnesium smelting process, we attempt, in this paper, feature exploration from both visual and current information through the cross-modal bridge. Inspired by the scaling success of Transformer models in natural language processing~\cite{Vaswani2017AttentionIA,OpenAI2023GPT4TR} and computer vision~\cite{Dosovitskiy2020An,Ni2022ExpandingLP,Driess2023PaLMEAE,Wang2023ActionCLIP}, we propose a novel cross-modal Transformer, dubbed FmFormer, to explore both spatial (visual) and temporal (visual and current) features, as illustrated in Fig.~\ref{fig:FMF}(b). First, we present a novel multiscale tokenization paradigm that effectively samples 3D patches with varying local receptive fields from visual modality (video) and 1D vectors from current modality, and embeds the samples into features (also known as tokens). Despite the considerably large dimensionality gap between 3D video and 1D current data, the tokenization paradigm seamlessly converts the multi-modal inputs into features with equivalent dimensions, enabling efficient exploration of intrinsically correlated features using succeeding attention mechanisms. More importantly, the proposed multiscale tokenization paradigm facilitates pixel-level anomaly detection, progressing from coarse to fine through hierarchical reassembly of the multiscale tokens. After tokenization, the proposed FmFormer processes the internal features of each modality separately through a self-attention. Subsequently, a bidirectional cross-attention is utilized to obtain correlation features in both current-to-visual direction and visual-to-current direction. By leveraging cross-modal learning, a multi-head decoder is tailored to process correlated tokens for class-level predictions and to assemble current-to-visual tokens for locating anomaly regions (pixel-level predictions), achieving accurate and robust anomaly detection.

To demonstrate the effectiveness of the proposed FmFormer for learning the cross-modal information, we present a novel cross-modal benchmark for the fused magnesium smelting process. We collected over 1,000 hours of synchronously acquired videos and three-phase alternating current data from different production batches, and selected over $2.2\times10^6$ samples to build the benchmark. By taking full advantage of the information from the two modalities, the proposed FmFormer is able to accurately detect the anomalies in fused magnesium smelting processes under extreme interferences, such as high frequency current fluctuation and visual occlusion caused by heavy water mist.

Summarized below are the main contributions of this paper:
\begin{itemize}
\item A novel cross-modal Transformer (FmFormer) is proposed with a cascading structure of alternately stacked self-attention and bidirectional cross-attention layers. It progressively encodes internal features of each modality and correlation features across modalities, fully leveraging the strengths of different modalities to achieve accurate and robust anomaly detection in the context of the fused magnesium smelting process.
\item A novel multiscale tokenization paradigm is tailored for generating token sets with varying local receptive fields. This paradigm facilitates the hierarchical reconstruction of pixel-level anomaly predictions with high localization accuracy.
\item A multi-head decoder is designed to translate correlated tokens into class-level and pixel-level predictions, enabling concurrent anomaly detection and anomaly localization.
\item A pioneering cross-modal benchmark with over 2.2 million samples of synchronously acquired video and current for anomaly detection in a real industrial scenario. To the best of our knowledge, the presented benchmark is the first cross-modal benchmark for anomaly detection of fused magnesium smelting processes. We are releasing the proposed benchmark at \url{https://github.com/GaochangWu/FMF-Benchmark} to foster research on cross-modal learning for industrial scenarios.
\end{itemize}


\section{Related Work}\label{sec:relatedwork}
\subsection{Learning-based Anomaly Detection}
Anomaly detection is defined as discovering patterns in data that do not match desired behavior, which can be mainly categorized into learning-based, statistical-based, information theory-based and spectral theory-based~\cite{Chandola2009AnomalyDA}. In this paper, we only focus on learning-based anomaly detection approaches. 
For time-series inputs, Yin \textit{et al.}~\cite{Yin2022AnomalyDB} converted the one-dimensional data into two-dimensional data using a sliding window scheme and implemented anomaly detection by combining convolutional layers and LSTM cells into an autoencoder architecture.
Barz \textit{et al.}~\cite{Barz2018DetectingRO} introduced a maximizes divergent intervals framework for spatio-temporal anomaly detection in an unsupervised learning manner. 
To achieve unsupervised learning from only normal data, Liu \textit{et al.}~\cite{Liu2023Time} introduced generative adversarial networks that learn to reconstruct the time-series signals in a low-dimensional space. When the reconstruction error is large, the input signal is considered to be abnormal. Due to the effective reduction of annotation cost, this unsupervised learning style has also been extended to image and video anomaly detection. However, significant noise and occlusion interference in harsh industrial environments greatly degrade model performance.
For video anomaly detection with supervised learning, Zaheer \textit{et al.}~\cite{Zaheer2022GenerativeCL} introduced a generative adversarial learning method using 3D convolutional backbones. But convolution operation has a limited receptive field, resulting in biased local information.

To solve the local-bias problem mentioned above, the self-attention mechanism in Transformer~\cite{Vaswani2017AttentionIA} is designed to compute the correlation of each element to all the other elements, resulting in a global perception.
Based on this structure, Xu \textit{et al.}~\cite{Xu2022Anomaly} introduced an anomaly Transformer that uniformly models both local and global information, which is challenging for convolution operations. 
Chen \textit{et al.}~\cite{Chen2023BidirectionalSA} proposed a bidirectional spatio-temporal Transformer to predict urban traffic flow using graph-based traffic representation. Cross-attentions of past-to-present and future-to-present directions are designed to learn the temporal tendency from the traffic data.
To solve high-dimensional vision tasks, Dosovitskiy \textit{et al.}~\cite{Dosovitskiy2020An} extended Transformer by decomposing an image into a sequence of vectorized patches and converting them into token representations. This extended version is called ViT.
For visual anomaly detection using 3D video input, Li \textit{et al.}~\cite{Li2022SelfTrainingML} applied a convolutional Transformer to encode 2D image slices into feature vectors, and then used another convolutional Transformer to decode these vectors into detection result.

Recently, a growing number of studies have shown that pure Transformer structures are also capable of learning representative features from high-dimensional video without using convolution-based encoders. For instance, Arnab \textit{et al.}~\cite{Arnab2021ViViTAV} extended the 2D tokenization of ViT~\cite{Dosovitskiy2020An} into 3D space via 3D convolution, called tubelet embedding. 3D patches are extracted from the input video to construct vectorized tokens, which are further fed into a standard Transformer. Piergiovanni \textit{et al.}~\cite{Piergiovanni2023RethinkingVV} generalized this idea and proposed to utilize tubes of different shapes to sparsely sample the input video. Since 3D patches and 2D patches are jointly extracted in the tokenization, both 3D videos and 2D images can be seamlessly applied for network training. Different from the aforementioned tokenization paradigms that samples 3D patches from the input video, we highlight the idea of dilated convolution~\cite{Yu2015Dilatconv} and propose a novel dilated tokenization to construct tokens with a larger local receptive field. This tokenization paradigm naturally forms a multiscale mechanism for efficiently reconstructing pixel-level anomaly detection.

Despite the powerful modeling capabilities of Transformers, models \revise{using unimodal input~\cite{wu2019abnormal,Lu2022SemiSupervisedCM,Liu2023DisturbanceRA,10587314} in complex industrial environments remain limited}. This limitation motivates us to combine both current and visual modalities for anomaly detection, leveraging their complementary strengths to enhance detection accuracy and robustness.

\subsection{Cross-Modal Learning}
Human-beings inherently have the ability to perceive and process cross-modal information such as language, sound, image, etc.
With the rapid increase in the diversity of acquired information and the improvement of computing power, cross-modal learning is becoming an emerging direction in the field of artificial intelligence, e.g., hybrid imaging~\cite{Zhao2018CrossScaleRL,Xu2020U2FusionAU,Zhou2021CrossMPICS,Shao2021LocalTransAM,10163247}, visual question answering~\cite{Benyounes2017MUTANMT,Ju2021PromptingVM,Wu2022BidirectionalCK}, visual-text retrieval~\cite{10057259,10445315,10345597}, robot perception~\cite{Yang2021LearningVQ} and chatbot (the ChatGPT~\cite{Brown2020LanguageMA,OpenAI2023GPT4TR}). 

A straightforward multi-modal learning scheme is to process each modality input with a different branch of network and then merge them to generate the fused feature. 
For example, Zhou \textit{et al.}\cite{Zhou2022IdentificationOA} introduced a multi-source information fusion method that employs a CNN-based image recognition branch and a current processing branch, with detection results individually predicted and then fused using a support vector machine. 
Similarly, Bu \textit{et al.}\cite{Bu2023ProcessOP} used a CNN and a Multi-Layer Perceptron (MLP) to extract image and current features, respectively, for anomaly prediction. However, the above linear fusion scheme neglects the correlation between the modalities, resulting in the underutilization of multi-modal information.
The key difference between the proposed FmFormer and the aforementioned studies is that we explicitly model the information interaction between video and current through cross-attention, which explores the bidirectional correlation between these two modalities, i.e., cross-modal learning~\cite{Xu2023MultimodalLW}.

\begin{figure*}
	\begin{center}
		\includegraphics[width=0.99\linewidth]{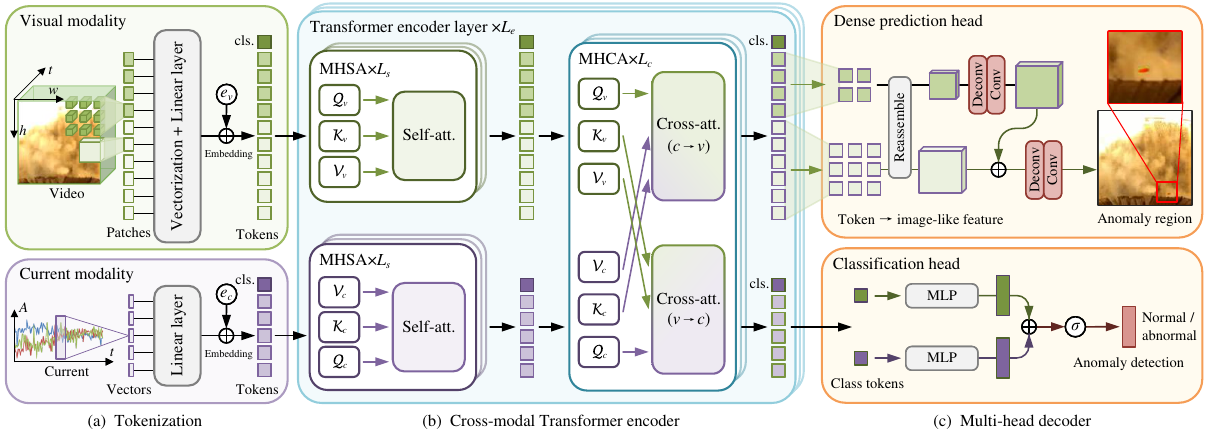}
	\end{center}
 \vspace{-3mm}
	\caption{An overview of the proposed FmFormer for anomaly detection in fused magnesium smelting processes: (a) The tokenization converts 3D video and 1D current into sets of vectors, or tokens, embedded with positional information. We introduce a novel multiscale video tokenization paradigm based on dilated sampling, which generates tokens of varying spatial scales while keeping consistent dimensions. (b) The cross-modal Transformer encoder learns representations from video and current tokens by exploring internal features with self-attention and correlation features through bidirectional cross-attention, producing current-to-visual and visual-to-current tokens. (c) A multi-head decoder is designed for dense and class prediction of abnormal conditions. The dense prediction head reassembles multi-scale video tokens (in light green) into image-like feature representations, while the classification head integrates class tokens from both modalities (in dark green and dark violet) into a classification vector.}
	\label{fig:pipeline}
\vspace{-3mm}
\end{figure*}

Explicit cross-modal learning involves modeling the correlation between two modalities in the embedding space. For example, Ben-younes \textit{et al.}~\cite{Benyounes2017MUTANMT} model the visual question answering task as a bilinear interaction between visual and linguistic features, and introduced a Tucker decomposition of the correlation tensor to explicitly control the model complexity.
\revise{Park~\textit{et al.}~\cite{10163247} developed spatial and channel Transformers to facilitate cross-modal learning and fusion of infrared and visible images with the same dimensions.
To reduce the dimensionality gap between 3D video and 1D text, pre-trained CNN and RNN backbones are commonly used to align the dimensions across modalities, with modality interaction implemented either through cross-attention~\cite{10057259} or self-attention with concatenated features~\cite{Lei2021LessIM}.}
Instead of compressing visual features, Yang \textit{et al.}~\cite{yang2022zero} proposed to embed each video frame into a visual token via a well-pre-trained visual-language model~\cite{Radford2021LearningTV}.
Wu \textit{et al.}~\cite{Wu2022RevisitingCT,Wu2023TransferringVM} revisited the video classification task and converted it to a cross-modal learning problem by constructing a correlation matrix for video and language embeddings. This video classification paradigm is able to leverage well-pre-trained language models to generate precise semantic information.
Although the spatio-temporal correlation is explored in these studies, each sample (e.g., video frame) in the time dimension is treated equally. Wu \textit{et al.}~\cite{Wu2022BidirectionalCK} further introduced a temporal concept spotting mechanism using video and language embeddings to produce a category dependent temporal saliency map, which helps to enhance video recognition. 

Motivated by these studies, we employ bidirectional cross-attention to explicitly explore correlated features across modalities.
The main differences between our FmFormer and existing cross-modal approaches are: i) We use a cascading structure to alternatively stack self-attention and cross-attention, enabling the network to progressively capture both internal features within each modality and correlations across modalities; ii) Our encoder processes visual tokens at multiple scales through a multiscale tokenization module, addressing the dimensionality gap between the modalities while enhancing robustness to visual interferences; iii) Our multi-head decoder reassembles multiscale visual tokens into image-like feature representations for hierarchical feature fusion, simultaneously tackling class-level and pixel-level prediction tasks for improved anomaly detection and localization accuracy.

\section{FmFormer for Anomaly Detection}\label{sec:fmformer}
The proposed FMF Transformer (FmFormer) has two sources of inputs, video with two spatial dimensions and one temporal dimension and three-phase alternating current with one temporal dimension. The FmFormer is composed of a tokenization module, a cross-modal Transformer encoder, and a multi-head decoder. The tokenization module converts 3D video and 1D current into two sets of 1D features. The encoder then explores the correlation between two modalities by using these features. The decoder consists of two functionally different parts, a classification head for class-level anomaly detection and a dense prediction head for spotting anomaly regions, i.e., pixel-level detection. The overall architecture of the proposed FmFormer is illustrated in Fig.~\ref{fig:pipeline}.

\subsection{Tokenization}
We prepare cross-modal features for our FmFormer by converting the high dimensional video into a set of vectorized patches and the low dimensional current into a set of vectors. Since they are one-dimensional vectors, we will call them ``tokens''~\cite{Vaswani2017AttentionIA,Dosovitskiy2020An,Arnab2021ViViTAV} throughout the rest of this paper. Fig.~\ref{fig:pipeline}(a) illustrates the overall pipeline of the tokenization.

\subsubsection{Video tokenization}
Consider a video clip (visual input) $x_v\in\mathbb{R}^{T_v\times H\times W\times D_v}$, where $T_v$ is the number of frames, $H$ and $W$ are the pixel numbers of each frame in height and width (i.e., image resolution), and $D_v=3$ represents the RGB color space. Standard video tokenization~\cite{Arnab2021ViViTAV,Neimark2021VideoTN} extends the 2D embedding for image~\cite{Dosovitskiy2020An} to 3D space via extracting non-overlapping spatio-temporal patches. Specifically, for a patch $p_v\in\mathbb{R}^{t_v\times h_v\times w_v\times D_v}$, the standard tokenization produces a patch set $p_v\in P_v$ ($\vert P_v \vert=n_t\times n_h\times n_w$) from the 3D video, where $n_t=\lfloor\frac{T_v}{t_v}\rfloor$, $n_h=\lfloor\frac{H}{h_v}\rfloor$, $n_w=\lfloor\frac{W}{w_v}\rfloor$, and $\lfloor\cdot\rfloor$ denotes round down.

In addition to the standard tokenization described above, we propose a novel dilated tokenization that is capable of constructing a multiscale token set. Similar to the dilated convolution~\cite{Yu2015Dilatconv}, we sample the element in the input video sparsely in the spatial dimensions, producing patches of the same size but with a larger local receptive field (as shown in Fig.~\ref{fig:pipeline}(a)). For a patch $p_v^d$ of the same size of $p_v$, the dilated tokenization generates a smaller patch set $p_v^d\in P_v^d$ ($\vert P_v^d\vert=n_t\times n_h^d\times n_w^d$), where $n_h^d=\lfloor\frac{H}{(h_v-1)\times d+1}\rfloor$, $n_w^d=\lfloor\frac{W}{(w_v-1)\times d+1}\rfloor$, and $d$ is the dilation rate.

The integration of standard tokenization and dilated tokenization constructs a multiscale video patch set. Each 3D patches in $P_v$ and $P_v^d$ are then flattened as 1D vectors of size $1\times (t_v\cdot h_v\cdot w_v\cdot D_v)$ and mapped into to tokens of the desired dimension $1\times D$. 
Following the vanilla Transformer~\cite{Vaswani2017AttentionIA}, we prepend a learnable class token and then add positional embeddings to retain positional information. The proposed multiscale tokenization module can be formulated as:
\begin{equation}\label{eq:video_token}
z_v=\{z_v^{cls}, \{\phi_v(P^d_v), \phi_v(P_v)\} \mathcal{W}_v^z\} + e_v,
\end{equation}
where $\phi_v(\cdot)$ indicates the vectorization operation, $\{\cdot, \cdot\}$ represents the concatenation of the tokens along the first dimension, $\mathcal{W}^z_v\in \mathbb{R}^{(t_v\cdot h_v\cdot w_v\cdot D_v)\times D}$ is the weight of the linear mapping layer, $z_v^{cls}\in\mathbb{R}^{1\times D}$ is the concatenated video class token, and $e_v\in\mathbb{R}^{(n_t\cdot n_h^d\cdot n_w^d+n_t\cdot n_h\cdot n_w+1)\times D}$ is the positional embeddings. For simplicity, we denote the dimension of the resulting video tokens $z_v\in\mathbb{R}^{N_v\times D}$, where $N_v = n_t\times n_h^d\times n_w^d+n_t\times n_h\times n_w+1$. Instead of using fixed sine/cosine embeddings in the vanilla Transformer~\cite{Vaswani2017AttentionIA}, we adopt learnable positional embeddings $e_v$ as the vision Transformers, which has been used in different vision tasks~\cite{Dosovitskiy2020An,Ranftl2021VisionTF}.

\subsubsection{Current tokenization}
The current sequence is denoted as $x_c\in\mathbb{R}^{T_c\times D_c}$, where $T_c$ is the sequence length of the input current and $D_c=3$ represents the three phases. In fact, the current sequence itself is an excellent set of tokens, we just need to map them into tokens $z_c$ of the desired dimension for the following Transformer layers:
\begin{equation}\label{eq:token}
z_c=\{z_c^{cls}, x_c \mathcal{W}_c^z\} + e_c,
\end{equation}
where $\mathcal{W}^z_c\in \mathbb{R}^{D_c\times D}$ is the weight matrix of a linear mapping layer, $z^{cls}_c\in\mathbb{R}^{1\times D}$ is the concatenated current class token, and $e_c\in \mathbb{R}^{(T_c+1)\times D}$ is learnable positional embeddings. We denote the dimension of the current tokens $z_c\in\mathbb{R}^{N_c\times D}$, where $N_c = T_c+1$.

Despite the significant dimensionality gap between 3D video and 1D current, the designed tokenization paradigm is able to seamlessly convert this dimension difference to the quantity difference of tokens, while keeping any two tokens with matching dimensions. This characteristic of the tokenization contributes to elegantly migrating the attention mechanism to cross-modal learning.

\subsection{Cross-Modal Transformer Encoder}
In the Transformer encoder, self-attention and bidirectional cross-attention are applied to encode the internal features and correlation features using video tokens $z_v$ and current tokens $z_c$. To better learn anomaly representation across modalities, we alternately cascade multiple layers of self-attention and cross-attention, as illustrated in Fig.~\ref{fig:pipeline}(b).

\subsubsection{Internal feature encoding using self-attention}
The tokenized representations of video and current inputs are barely aligned in the embedding space. In other words, it is impossible to find any correlation between the video tokens $z_v$ and the current tokens $z_c$. Therefore, before calculating the correlation we use self-attention mechanism to obtain the internal features of video and current tokens and roughly align them through end-to-end training.

We use Multi-Head Self-Attention (MHSA)~\cite{Vaswani2017AttentionIA} to process the video tokens $z_v$ and the current tokens $z_c$, respectively. First, the video and current tokens are mapped into queries $\mathcal{Q}$, keys $\mathcal{K}$, and values $\mathcal{V}$ with dimension $D$:
\begin{equation}\label{eq:qkv}
\mathcal{Q}_f = \phi_{LN}(z_f)\mathcal{W}^\mathcal{Q}_f,\mathcal{K}_f=\phi_{LN}(z_f)\mathcal{W}^\mathcal{K}_f,\mathcal{V}_f=\phi_{LN}(z_f)\mathcal{W}^\mathcal{V}_f,
\end{equation}
with $f=v$ for video and $f=c$ for current, $\mathcal{W}^\mathcal{Q}_f,\mathcal{W}^\mathcal{K}_f,\mathcal{W}^\mathcal{V}_f\in\mathbb{R}^{D\times D}$ stand for the weight matrices of linear mapping layers, and $\phi_{LN}$ indicates the layer normalization~\cite{Ba2016LayerN}. Then the self-attention is formulated as:
\begin{equation}
A_f^{(l_s)}=\phi_A(\mathcal{Q}_f^{(l_s)}, \mathcal{K}_f^{(l_s)}, \mathcal{V}_f^{(l_s)}) = \sigma(\frac{\mathcal{Q}_f^{(l_s)} (\mathcal{K}_f^{(l_s)})^T}{\sqrt{D/L_s}})\mathcal{V}_f^{(l_s)},
\end{equation}
where $A_f^{(l_s)}$ denotes one of the attention heads, $l_s\in \{1,\cdots,L_s\}$ with $L_s$ being the number of attention in the MHSA, $\phi_A$ stands for the attention mechanism, and $\sigma$ denotes the softmax non-linearity. Here, $\mathcal{Q}_f$, $\mathcal{K}_f$ and $\mathcal{V}_f$ are evenly divided into $\mathcal{Q}^{(l_s)}_f,\mathcal{K}^{(l_s)}_f,\mathcal{V}^{(l_s)}_f\in\mathbb{R}^{N_f\times \frac{D}{L_s}}$ for the computation of each self-attention head.
To jointly learn different representation subspaces from training instances, the MHSA concatenates multiple self-attention results as follows:
\begin{align}\label{eq:MHSA1}
\mathcal{A}_f=\{(A_f^{(1)})^T,\dots,(A_f^{(L_s)})^T\}^T\mathcal{W}_f^{s}+z_f,
\end{align}
where $\mathcal{W}_f^{s}\in\mathbb{R}^{D\times D}$ stands for a weight matrix to linearly map the concatenated self-attention features into dimension $D$, and $\mathcal{A}_f\in\mathbb{R}^{N_f\times D}$ is the resulting MHSA feature. The final tokens processed by the MHSA is:
\begin{align}\label{eq:MHSA2}
z_f \leftarrow \phi_{MLP}(\mathcal{A}_f)+\mathcal{A}_f,
\end{align}
where $\phi_{MLP}$ denotes the MLP with two linear layers.

Explicitly interacting across tokens makes self-attention inherently a global operation. Therefore, the Transformer encoder is superior in the capability to capture the fine-grained features in the spatio-temporal dimensions of each modality compared to other backbones such as CNN or LSTM.

\subsubsection{Correlation feature encoding using bidirectional cross-attention}
To facilitate feature exploration across different modalities, we utilize the cross-attention between visual and current modalities in a bidirectional manner. The implemented cross-attention has the following characteristics: i) Each token in one modality can interact with all the tokens in the other modality; ii) The resulting attention map has two directions, current-to-visual and visual-to-current, and is directional sensitive; iii) The dimension of each modality remain unchanged.

First, the video and current tokens processed by the self-attention, $z_v$ and $z_c$, are mapped into $\mathcal{Q}_v,\mathcal{K}_v,\mathcal{V}_v\in\mathbb{R}^{N_v\times D}$ and $\mathcal{Q}_c,\mathcal{K}_c,\mathcal{V}_c\in\mathbb{R}^{N_c\times D}$ using (\ref{eq:qkv}), respectively. Inspired by the image-language cross-modal learning in~\cite{Li2021GroundedLP}, we define the cross-attention in a bidirectional manner:
\begin{align}
\begin{split}\label{eq:cross-att}
A_{c\rightarrow v}^{(l_c)} &= \phi_A(\mathcal{Q}^{(l_c)}_v, \mathcal{K}^{(l_c)}_c, \mathcal{V}^{(l_c)}_c)=\sigma(\frac{\mathcal{Q}^{(l_c)}_v (\mathcal{K}^{(l_c)}_c)^T}{\sqrt{D/L_c}})\mathcal{V}^{(l_c)}_c,\\
A_{v\rightarrow c}^{(l_c)} &= \phi_A(\mathcal{Q}^{(l_c)}_c, \mathcal{K}^{(l_c)}_v, \mathcal{V}^{(l_c)}_v)=\sigma(\frac{\mathcal{Q}^{(l_c)}_c (\mathcal{K}^{(l_c)}_v)^T}{\sqrt{D/L_c}})\mathcal{V}^{(l_c)}_v,\\
\end{split}
\end{align}
where $A^{(l_c)}_{c\rightarrow v}\in\mathbb{R}^{N_v\times \frac{D}{L_c}}$ is the $l_c$th cross-attention head in current-to-visual direction, and $A^{(l_c)}_{v\rightarrow c}\in\mathbb{R}^{N_c\times \frac{D}{L_c}}$ is the $l_c$th cross-attention head in visual-to-current direction. Similar to the MHSA, the Multi-Head Cross-Attention (MHCA) with the number of heads $L_c$ is also implemented using (\ref{eq:MHSA1}) to generate bidirectional MHCA $\mathcal{A}_{c\rightarrow v}\in\mathbb{R}^{N_v\times D}$ and $\mathcal{A}_{v\rightarrow c}\in\mathbb{R}^{N_c\times D}$. Then the resulting tokens encoded with cross-modal information are given as:
\begin{align}\label{eq:MHCA2}
\begin{split}
z_{c\rightarrow v} &\leftarrow \phi_{MLP}(\mathcal{A}_{c\rightarrow v})+\mathcal{A}_{c\rightarrow v},\\
z_{v\rightarrow c} &\leftarrow \phi_{MLP}(\mathcal{A}_{v\rightarrow c})+\mathcal{A}_{v\rightarrow c},
\end{split}
\end{align}
where $z_{c\rightarrow v}$ is the current-to-visual tokens and $z_{v\rightarrow c}$ is the visual-to-current tokens.

It should be noted that the MHCA in (\ref{eq:cross-att}) is directional sensitive, i.e., the attention maps $\sigma(\frac{\mathcal{Q}^{(l_c)}_v (\mathcal{K}^{(l_c)}_c)^T}{\sqrt{D/L_c}})\in\mathbb{R}^{N_v\times N_c}$ and $\sigma(\frac{\mathcal{Q}^{(l_c)}_c (\mathcal{K}^{(l_c)}_v)^T}{\sqrt{D/L_c}})\in\mathbb{R}^{N_c\times N_v}$ hold different meanings based on the direction of the information flow between the visual and current modalities. Specifically, in current-to-visual directional MHCA, the visual features query the current features, producing an attention map that encodes the correlation of each visual token according to its attention from all current tokens, and vice versa. Therefore, the bidirectional cross-attention enables the proposed FmFormer to achieve global perception across modalities. This also ensures that the model captures more comprehensive features, enabling accurate and robust anomaly detection in the fused magnesium smelting process, particularly when one of the modality are less reliable.

\subsection{Multi-Head Decoder}
Since the processed tokens can be categorized into class tokens and regular tokens as described in (\ref{eq:video_token}) and (\ref{eq:token}), we can naturally perform multiple types of anomaly detection tasks, e.g., dense prediction (pixel-level anomaly detection) and class prediction (class-level anomaly detection), through a multi-head decoder setting, as shown in Fig.~\ref{fig:pipeline}(c).

\subsubsection{Dense prediction head}\label{sec:densehead}
The dense prediction head reassembles the multiscale video tokens into image-like feature representations~\cite{Ranftl2021VisionTF}, which are then progressively fused for spotting anomaly regions. Since the information from the current tokens and the class tokens has been integrated into the current-to-visual tokens after the Transformer encoder, we only use the regular current-to-visual tokens (i.e., ignoring the class token marked in dark green in Fig.~\ref{fig:pipeline}(b)) for the prediction.

Let $z_{c\rightarrow v}=\{z_{c\rightarrow v}^{cls},z^{r,d}_{c\rightarrow v},z_{c\rightarrow v}^r\}$, where $z^{r,d}_{c\rightarrow v}\in\mathbb{R}^{(n_t\cdot n_h^d\cdot n_w^d)\times D}$ and $z^{r}_{c\rightarrow v}\in\mathbb{R}^{(n_t\cdot n_h\cdot n_w)\times D}$ denote the tokens stemmed from the dilated tokenization and the standard tokenization, respectively (as shown in Fig.~\ref{fig:pipeline}(c)).
First, we reassemble the multiscale video tokens by reshaping $z^{r,d}_{c\rightarrow v}$ and $z^{r}_{c\rightarrow v}$ into 3D features of sizes $n_h^d\times n_w^d\times (n_t\cdot D)$ and $n_h\times n_w\times (n_t\cdot D)$ and squeezing the channel numbers with $1\times 1$ convolutions:
\begin{align}\label{eq:assemble}
\begin{split}
I_v^d &= \phi_{conv1\times1}(\phi_{r}(z^{r,d}_{c\rightarrow v})),\\
I_v &= \phi_{conv1\times1}(\phi_{r}(z^{r}_{c\rightarrow v})),
\end{split}
\end{align}
where $\phi_{r}$ is the reshape operation, $\phi_{conv1\times1}$ denotes the $1\times 1$ convolution, and $I^d_v\in\mathbb{R}^{n_h^d\times n_w^d\times D}$ and $I_v\in\mathbb{R}^{n_h\times n_w\times D}$ are the resulting image-like features.

Note that the spatial resolutions of the two features are different, representing visual features at different local receptive field scales. We therefore blend the two features by first upsampling $I^d_v$ with a sequential stack of deconvolution (also known as transposed convolution) and convolution, and then adding them together:
\begin{equation}\label{eq:fusion}
I^b_v = I_v + \phi_{conv}(\phi_{deconv}(I^d_v)),
\end{equation}
where $\phi_{deconv}$ denotes the deconvolution and $I^b_v\in\mathbb{R}^{n_h\times n_w\times D}$ is the blended feature. The final dense prediction of the anomaly regions is generated by reconstructing the blended feature $I^b_v$ to the desired spatial resolution $H\times W$ with several sequential stacks of transposed convolution and convolution:
\begin{equation}\label{eq:dense_pred}
\hat{y}^{pix} = \phi_{conv_{1\times1}}(\phi_{conv}(\phi_{deconv}(\cdots\phi_{conv}(\phi_{deconv}(I^b_v))\cdots))),
\end{equation}
where $\hat{y}^{pix}\in\mathbb{R}^{H\times W\times K}$ denotes the pixel-level prediction result and $K=2$ indicates the binary classification task (normal and abnormal).

\subsubsection{Classification head}
The classification head transforms the class tokens into an anomaly prediction, i.e., a classification vector. First, we map the class tokens (marked in dark green and dark violet in Fig.~\ref{fig:pipeline}(c)) processed by the Transformer encoder, $z_{c\rightarrow v}^{cls}\in\mathbb{R}^{1\times D}$ and $z_{v\rightarrow c}^{cls}\in\mathbb{R}^{1\times D}$, to vectors $\hat{y}_{c\rightarrow v}^{cls}\in\mathbb{R}^{1\times K}$ and $\hat{y}_{v\rightarrow c}^{cls}\in\mathbb{R}^{1\times K}$ using two small MLPs. 
Then the detection result is obtained via a simple Linear Fusion (LF) mechanism formulated as follows:
\begin{equation}\label{eq:late_fusion}
\hat{y}^{cls}=\sigma(\hat{y}_{c\rightarrow v}^{cls} + \hat{y}_{v\rightarrow c}^{cls}),
\end{equation}
where $\sigma$ denotes the softmax non-linearity and $\hat{y}^{cls}\in\mathbb{R}^{1\times K}$ indicates the anomaly detection result.

\section{Benchmark and Network Training}\label{sec:benchmark_and_training}
\subsection{Fused Magnesium Smelting Process Benchmark}\label{sec:benchmark}
The benchmark contains cross-modal data from fused magnesium smelting processes for a total of 3 production batches, in which the video data is captured by industrial cameras, and the current data is sampled by PLC control systems, as shown on the right of Fig.~\ref{fig:FMF}(a). Rectangular wave signal generators are employed to trigger industrial cameras and PLC control systems to capture video frames and currents, enabling the synchronous acquisition of video frames and current measurements.
The raw data contains a total of more than 1,000 hours of synchronized videos and current sequences.
However, directly using the raw dataset as the benchmark causes class imbalance, as abnormal conditions are infrequent. Therefore, we carefully selected over $2.2\times 10^6$ samples (about 25 hours, 25 samples per second, and spatial resolution of $1440\times2560$ for each raw video frame) through the entire raw dataset to keep the ratio of normal to abnormal samples close to $1:1$. The benchmark comprises two types of datasets: a pixel-level annotated dataset with approximately $2.7\times 10^5$ samples and a class-level annotated dataset with around $2.0\times10^6$ samples. More detailed statistics for the benchmark are listed in Table~\ref{table:benchmark}. The fused magnesium smelting process benchmark will be available at \url{https://github.com/GaochangWu/FMF-Benchmark}.

\begin{table}
\caption{Statistics of the fused magnesium smelting process benchmark.}
\vspace{-3mm}
\label{table:benchmark}
\begin{center}
\begin{tabular}{l p{2.5cm}<{\centering} p{2.5cm}<{\centering} p{1.1cm}<{\centering}}
\toprule
\multirow{2}*{Dataset} & Pixel-level & Class-level & \multirow{2}*{Total}\\
                       & annotated dataset & annotated dataset & \\
\hline
Normal & 129,859 &  940,113 & 1,069,972\\
Abnormal & 140,062 &  1,026,663 & 1,166,725\\
Total &  269,921 & 1,966,776 &  2,236,697 \\
\bottomrule
\vspace{-5mm}
\end{tabular}
\end{center}
\end{table}

\begin{figure}
	\begin{center}
		\includegraphics[width=0.98\linewidth]{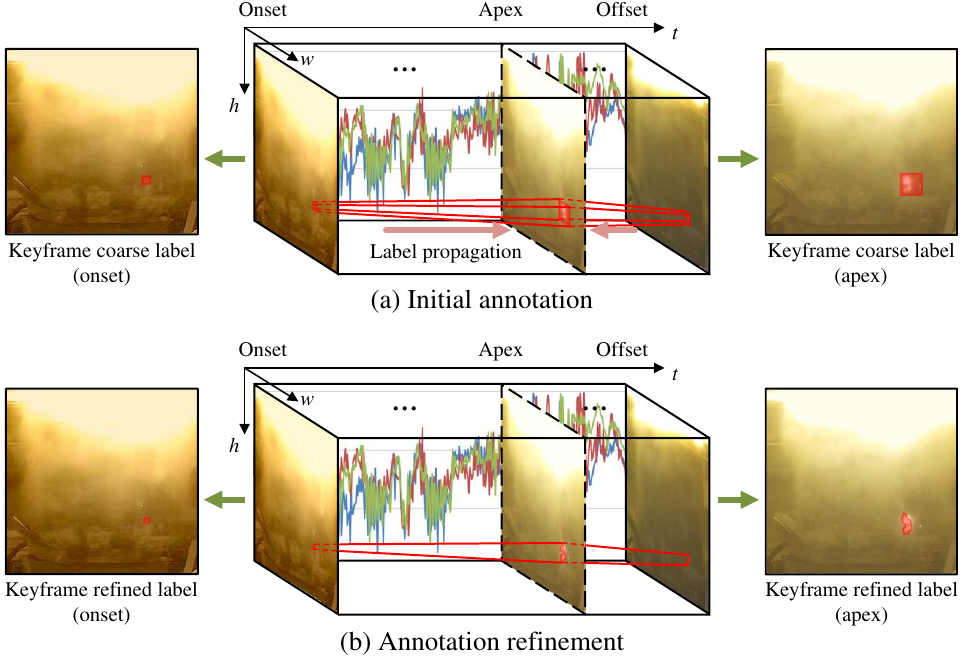}
	\end{center}
\vspace{-4mm}
	\caption{Efficient data annotation based on the piecewise temporal consistency of abnormal conditions in fused magnesium smelting processes. (a) The initial annotation is achieved by the following steps: i) determining the onset, apex and offset times of anomalies; ii) labelling the sparse set of frames (keyframes) with bounding boxes; iii) propagating boxes to other frames. (b) The annotation refinement is implemented to provide a more accurate representation of the anomaly region.}
	\label{fig:annotation}
\vspace{-3mm}
\end{figure}

\subsubsection{Data annotation}
The annotation of the large volume cross-modal data requires a mass of manpower. To accelerate the data annotation, we take advantage of the piecewise temporal consistency of abnormal conditions in fused magnesium smelting processes. Specifically, despite the data can be disturbed by noise or visual occlusion, the actual anomaly remains constant for a certain period of time. Based on this observation, we first determine the onset (starting), apex (highest intensity) and offset (ending) times of anomalies according to the visual and current features summarized by domain experts, and coarsely annotate a sparse set of frames using bounding boxes, each of which is called a keyframe. A more detailed description of the normal/abnormal current features can be referred in~\cite{Wu2015DataDrivenAC,Chai2017ACB,Wu2018NonlinearCT}. 
Next, a label propagation process is employed to assign a coarse label to every frame, as shown in Fig.~\ref{fig:annotation}(a). It linearly projects the bounding boxes to each frame between onset keyframe to apex keyframe, and then to each frame between apex keyframe and offset keyframe. To obtain more accurate labels that better fit the shape of the anomalies for dense prediction, we finally implement an annotation refinement process using a weighted median filter~\cite{WMF2014}, as shown in Fig.~\ref{fig:annotation}(b). It refines each label by utilizing the corresponding video frame as guidance. With the help of the efficient data annotation, the presented benchmark provides approximately $2.7\times 10^5$ cross-modal samples with pixel-level labels. 

\subsubsection{Analysis of data modality}
\begin{figure}
	\begin{center}
		\includegraphics[width=0.98\linewidth]{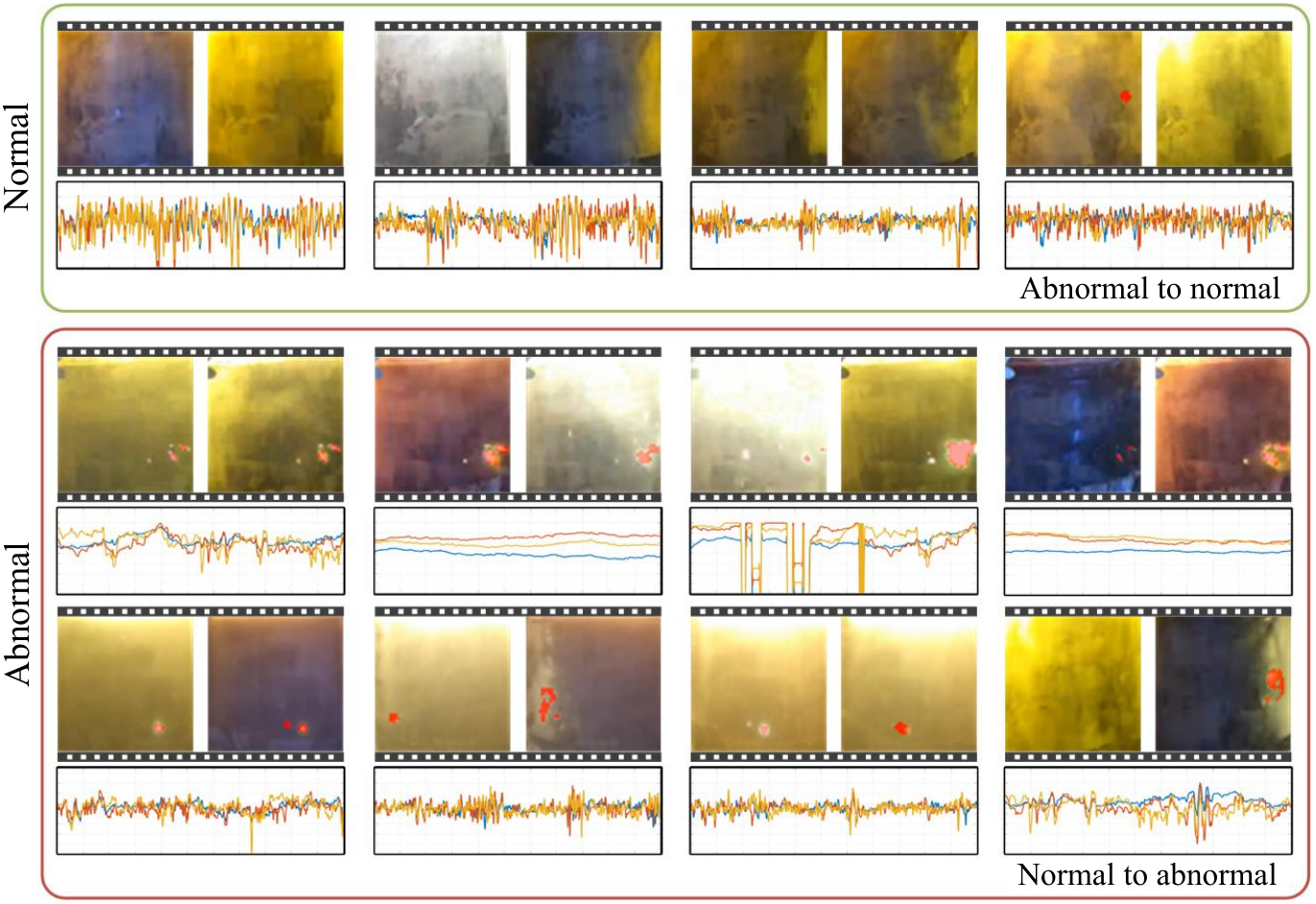}
	\end{center}
\vspace{-4mm}
	\caption{Examples of the fused magnesium smelting process benchmark. Each example demonstrates two frames (superimposed with the corresponding pixel-level labels) of a video clip and three-phase alternating current curve. In our cross-modal benchmark, we provide about 270,000 pixel-level annotated samples and 2.1 million class-level annotated samples.}
	\label{fig:dataset}
\vspace{-3mm}
\end{figure}

Fig.~\ref{fig:dataset} demonstrates examples of video frames (superimposed with labels) and current curves from the presented cross-modal benchmark. The first row illustrates four normal cases, while the last two rows show eight abnormal cases. From the first and last rows, it is evident that abnormal patterns in current signals can be easily obscured by data noise, complicating anomaly detection when relying solely on the current modality. In comparison, despite visual information showing salient features in most cases, it can be affected by occlusion from heavy water mist, as shown in the second and fourth cases of the last row. Additionally, weak feature anomalies can be obscured by intense flame light, as demonstrated by the second and third cases in the second row. Our benchmark provides both visual and current information to address these challenges and enhance the robustness of anomaly detection in fused magnesium smelting processes.

\subsection{Training Recipe}
Our training objective involves the minimization of a dense prediction loss and a classification loss:
$$
\arg\min_{\mathcal{W},e}\sum_{\langle x_v, x_c, y \rangle} \mathcal{L}^{pix}(\hat{y}^{pix}, y) + \alpha\mathcal{L}^{cls}(\hat{y}^{cls}, y),
$$
where $\mathcal{W}$ and $e$ are the learnable weights and learnable positional embeddings in the proposed FmFormer, $\langle x_v, x_c, y \rangle$ is the training set of input video $x_v\in\mathbb{R}^{T_v\times H\times W\times3}$, input current $x_c\in\mathbb{R}^{T_c\times3}$, and label $y\in\mathbb{R}^{H\times W\times K}$ triples, $\mathcal{L}^{pix}$ is the dense prediction loss term, $\mathcal{L}^{cls}$ is the classification loss term, and $\alpha$ is a hyperparameter to control the balance of the two terms. Specifically, we use the pixel-level cross-entropy loss for the dense prediction:
$$
\mathcal{L}^{pix}(\hat{y}^{pix}, y) = -\frac{1}{HW}\sum_{h,w}\sum_{k\in K} y_{(h,w,k)}\log \frac{\exp\hat{y}^{pix}_{(h,w,k)}}{\sum_{k'} \exp\hat{y}^{pix}_{(h,w,k')}},
$$
where $(h, w, k)$ indicates an element index in the prediction result $\hat{y}^{pix}$ or the pixel-label $y$. 
For the classification loss, we first employ an aggregation operation $\mathcal{G}$ to convert the pixel-level labels $y$ to the class-level label, simply by determining the presence of anomalies. Then we use class-wise cross-entropy loss for the classification:
\begin{align}\label{eq:loss_cls}
\mathcal{L}^{cls}(\hat{y}^{cls}, y) &= -\sum_{k\in K}\mathcal{G}(y;\tau)\log \hat{y}^{cls}_{(k)},
\end{align}
where $\mathcal{G}(\cdot;\tau)=[0, 1]$ when the number of anomaly pixels in the label $y$ exceeds $\tau$, and $\mathcal{G}(\cdot;\tau)=[1, 0]$ otherwise. $\tau$ is a threshold which we empirically set to $0.5$.

\subsection{Implementation Details}
\subsubsection{Architecture details}
\revise{Following typical Transformer backbones~\cite{Vaswani2017AttentionIA,Dosovitskiy2020An,Arnab2021ViViTAV}, we implement} four FmFormer models with varying capacities (Tiny, Small, Base, Large) with detailed configurations in Table~\ref{table:model_config}. In the table, ``Token dim.'' refers to the dimension of tokens, ``MLP dim.'' represents the dimension of the MLP in (\ref{eq:MHSA2}), ``MHSA heads \#'' indicates the number of attention heads ($L_s$ and $L_c$), and ``Encoder layers \#'' denotes the number of Transformer encoder layers. \revise{We maintain the same number of layers as in ViT~\cite{Dosovitskiy2020An} for each capacity configuration (noting that each layer contain one MHSA and one MHCA in our model), while adopting smaller token and MLP dimensions. This makes the model relatively lightweight compared to commonly used large-scale models~\cite{Dosovitskiy2020An,Arnab2021ViViTAV}, ensuring computationally efficient with satisfactory performance for industrial deployment, as analyzed in Sect.~\ref{sec:performanceVScapacity}.}

In the tokenization, we use a kernel of size $2\times8\times8$ (time $t_v$, height $h_v$, and width $h_w$) for the video tokenization. \revise{This setting is inspired by ViViT~\cite{Arnab2021ViViTAV}, with a smaller spatial size to accommodate our dilated tokenization, which incorporates a dilation rate of 2.}
In the dense prediction head, the $1\times1$ convolution layers in (\ref{eq:assemble}) have the same number of channels as the token dimension $D$, while each transposed convolution layers in (\ref{eq:fusion}) and (\ref{eq:dense_pred}) has a $2\times2$ kernel size with stride $2\times2$, \revise{progressively enhancing the spatial resolution of the feature maps. Following the implementation from~\cite{Ranftl2021VisionTF}}, each convolution layer in (\ref{eq:fusion}) and (\ref{eq:dense_pred}) is practically a sequential connection of a $3\times3$ convolution, a batch normalization, and a ReLU non-linearity. Due to the spatial reconstruction, we \revise{empirically} set the channel numbers of the transposed convolution layers in (\ref{eq:dense_pred}) as $[128, 64, 32, \cdots]$. \revise{To align with existing vision Transformers~\cite{Dosovitskiy2020An,Arnab2021ViViTAV}}, the MLPs in the MHSA, MHCA (in (\ref{eq:MHSA2}) and (\ref{eq:MHCA2})) and the classification head are sequential connections of a layer normalization, a linear layer, a GELU non-linearity, and another linear layer.

\subsubsection{Training details}
For the joint training of the dense prediction head and the classification head, we use the pixel-level annotated dataset as described in Section~\ref{sec:benchmark}. Note that pixel-level labels can also be converted into class-level labels. The dataset is divided into a training set with about $2.15\times10^5$ examples and a test dataset with $0.55\times10^5$ examples. In both the training and test datasets, the ratio of normal examples to abnormal examples remained close to $1:1$. We crop out areas of interest (i.e., furnaces) from the raw videos for the training and testing. The proposed FmFormer is implemented by using the Pytorch framework~\cite{Paszke2019PyTorchAI}. The AdamW solver~\cite{Loshchilov2017ADAMW} is applied as the optimization method, in which the batch size is set to 64. The step learning rate decay scheme is adopted with an initial learning rate of $5\times10^{-4}$, which then decays to $5\times10^{-5}$ after 20 epochs. The network (FmFormer-B) converges after 30 epochs of training, which takes about 5 hours on an NVIDIA TESLA V100.

\begin{table}[t]
\caption{Configuration for each FmFormer model.}
\vspace{-3mm}
\label{table:model_config}
\begin{center}
\begin{tabular}{l p{1.0cm}<{\centering} p{1.0cm}<{\centering} p{1.0cm}<{\centering} p{1.3cm}<{\centering} p{1.2cm}<{\centering}}
\toprule
Name & \makecell{Token\\ dim.} & \makecell{MLP\\ dim.}   & \makecell{MHSA\\ heads \#} & \makecell{Encoder\\ layers \#} & Param. \#\\
\hline
Tiny & 36 &  144 & 3 &  6 &  0.28M \\
Small &  48 &  192 & 3 &  6 &  0.48M \\
Base &  96 & 384 & 3 & 6 &  1.86M \\
Large &  96 & 384 & 3 & 12 & 3.64M \\
\bottomrule
\vspace{-6mm}
\end{tabular}
\end{center}
\end{table}

\section{Experiments}\label{sec:experiments}
In this section, we evaluate the proposed FmFormer (principally the base model) on the fused magnesium smelting process benchmark and compare it with several state-of-the-art learning-based anomaly detection methods that apply both unimodal inputs (current or visual) and cross-modal inputs. In the experiments, quantitative evaluations and visual comparisons specifically for dense prediction are performed. In addition, we empirically investigate the modules in the proposed FmFormer through several ablation studies. \revise{More applications to general industrial anomaly detection are explored in the supplementary material.}

\revise{We employ four commonly used metrics~\cite{Sokolova2009ASA}, accuracy, F1-score (F1), False Detection Rate (FDR), and Miss Detection Rate (MDR), for class-level anomaly detection. For pixel-level anomaly detection, we use Mean Intersection over Union (mIoU)~\cite{Everingham2010ThePV}.}
False Detection Rate (FDR), also known as false positive rate, is defined as the ratio of incorrectly predicted normal samples with respect to all the real normal (negative) samples.
Miss Detection Rate (MDR) is also known as false negative rate that indicates the ratio of incorrectly predicted abnormal samples to all the real abnormal (positive) samples.


\begin{table*}
\begin{threeparttable}
\caption{Quantitative comparison with state-of-the-art methods on the fused magnesium smelting process benchmark.}
\vspace{-3mm}
\label{table:exp1}
\begin{center}
\begin{tabular}{p{2.5cm} p{0.8cm}<{\centering} p{0.8cm}<{\centering} p{2.1cm}<{\centering} p{2.1cm}<{\centering} p{2.1cm}<{\centering} p{2.1cm}<{\centering} | p{2.1cm}<{\centering}}
\toprule
\multirow{2}*{Method} & \multicolumn{2}{c}{Modality}  & \multicolumn{4}{c|}{Classification} & Dense prediction\\
\cline{2-8}
& Visual & Current & $Acc \uparrow$ & $F1 \uparrow$ & $FDR \downarrow$  & $MDR \downarrow$ & $mIoU (\%)\uparrow$\\
\hline
Expert system~\cite{Wu2015DataDrivenAC} & \ding{55}  & \checkmark & 0.6851 & 0.7091 &  0.3586 &  0.1164  & -\\
Informer~\cite{Zhou2020InformerBE} & \ding{55}  & \checkmark & 0.7600 & 0.7960 &  0.3793 &  0.1164  & -\\
Flowformer~\cite{Wu2022FlowformerLT} & \ding{55}  & \checkmark & 0.7776   &  0.7950  &  0.2625  & 0.1866  & -\\
Flashformer~\cite{Dao2022FlashAttentionFA} & \ding{55}  & \checkmark & 0.7949   &  0.8065 &  0.2181  & 0.1935  & -\\
iTransformer~\cite{liu2023itransformer} & \ding{55}  & \checkmark & 0.7164 & 0.7305 &  0.2934 &  0.2748  & -\\
\textbf{FmFormer-B} (Ours) & \ding{55}  & \checkmark  & 0.7754 &  0.7688 & 0.1448 &  0.2954  & -\\
3DCRNN~\cite{Liu2023DisturbanceRA} & \checkmark & \ding{55} & 0.9690 & 0.9708 & 0.0380 & 0.0247  & 0.8289\\
ViT~\cite{Dosovitskiy2020An} & \checkmark & \ding{55} & 0.9650 & 0.9671 &  0.0408 &  0.0298  & 0.8124\\
ViViT~\cite{Arnab2021ViViTAV} & \checkmark & \ding{55} & 0.9687 & 0.9708 &  0.0466 &  0.0174  & 0.8321\\
TubeViT~\cite{Piergiovanni2023RethinkingVV} & \checkmark & \ding{55} & 0.9715 & 0.9732 &  0.0344 &  0.0231  & -\\
\textbf{FmFormer-B} (Ours) & \checkmark  & \ding{55} & 0.9731 &  0.9744 & 0.0224 &  0.0307  & \underline{0.8382}\\
DCNN-SVM~\cite{Zhou2022IdentificationOA} & \checkmark & \checkmark  & 0.8785 & 0.9086 &  0.2259 &  0.0282  & -\\
SSFGGAN~\cite{Bu2023ProcessOP} & \checkmark & \checkmark  &  0.8595 & 0.8967 &  0.2706 &  0.0245 & - \\
Unicoder-VL~\cite{Li2020UnicoderVLAU} & \checkmark & \checkmark  &  0.9680 & 0.9700 &  0.0428 &  0.0224 & 0.8185 \\
ClipBERT~\cite{Lei2021LessIM} & \checkmark & \checkmark  &  0.9757 & 0.9774 &  0.0340 & \underline{0.0157} & 0.8174 \\
CAPTURE~\cite{Zhan2021Product1MTW} & \checkmark & \checkmark  &  0.9744 & 0.9758 &  0.0260 & 0.0250 & 0.8312 \\
BiLM~\cite{yang2022zero} & \checkmark & \checkmark  &  \underline{0.9763} & \underline{0.9780} &  \textbf{0.0182} &  0.0274 & 0.8360 \\
\textbf{FmFormer-B} (Ours) & \checkmark & \checkmark  & \textbf{0.9837} & \textbf{0.9847} &  \underline{0.0217} &  \textbf{0.0117} &  \textbf{0.8409}\\
\bottomrule
\end{tabular}
\begin{tablenotes}
\footnotesize
\item[1] The best results are highlighted in \textbf{bold}, and the second-best results are \underline{underlined}.
\end{tablenotes}
\vspace{-3mm}
\end{center}
\end{threeparttable}
\end{table*}

\subsection{Comparison With State-of-the-Art Methods}
We evaluate the effectiveness of our base model, FmFormer-B, by comparing it with several baseline methods using unimodal settings (current/visual) and a cross-modal setting.

\subsubsection{Current modality}
Four state-of-the-art Transformer-based methods, including Informer~\cite{Zhou2020InformerBE}, Flowformer~\cite{Wu2022FlowformerLT}, Flashformer~\cite{Dao2022FlashAttentionFA} and iTransformer~\cite{liu2023itransformer}, are evaluated for current-based anomaly detection. 
Informer~\cite{Zhou2020InformerBE} employs a ProbSparse self-attention mechanism that generates sparse query-key pairs for efficient time-series modeling.
Flowformer~\cite{Wu2022FlowformerLT} linearizes Transformer free from specific inductive biases by applying the property of flow conservation into attention.
Flashformer~\cite{Dao2022FlashAttentionFA} uses tiling operation to reduce the number of GPU memory reads/writes to achieve IO-aware attention mechanism.
iTransformer~\cite{liu2023itransformer} embeds each series independently to the variate token instead of embedding temporal token in the vanilla Transformer.
In addition, a classical data-driven expert system method by Wu~\textit{et al.}~\cite{Wu2015DataDrivenAC} is also compared.

Table~\ref{table:exp1} presents the quantitative comparison with the baseline methods on the proposed fused magnesium smelting process benchmark. For current modal input, the proposed FmFormer-B achieves comparable performance with state-of-the-art Transformer-based methods that specifically designed for time-series prediction. Nevertheless, from the performance perspective, anomaly detection using pure current information is still far from practical industrial applications.

\subsubsection{Visual modality}
Four state-of-the-art methods or backbones are evaluated, which are 3DCRNN~\cite{Liu2023DisturbanceRA}, ViT~\cite{Dosovitskiy2020An}, ViViT~\cite{Arnab2021ViViTAV} and TubeVit~\cite{Piergiovanni2023RethinkingVV}. 3DCRNN~\cite{Liu2023DisturbanceRA} is a typical convolution recurrent-based anomaly detection framework specifically for fused magnesium smelting processes, utilizing a 2D convolutional LSTM to extract 3D (2D spatial and 1D temporal) features. ViT~\cite{Dosovitskiy2020An}, ViViT~\cite{Arnab2021ViViTAV} and TubeViT~\cite{Piergiovanni2023RethinkingVV} are three Transformer-based methods specifically designed for video input, and employ non-overlapping 2D patches~\cite{Dosovitskiy2020An}, 3D tubes~\cite{Arnab2021ViViTAV} and 3D tubes of different shapes~\cite{Piergiovanni2023RethinkingVV} for tokenization, respectively.

As shown in Table~\ref{table:exp1}, the visual models exploit visual features that are more stable and prominent than current information, which provide them a distinct advantage.
Besides, 3DCRNN~\cite{Liu2023DisturbanceRA} achieves comparable performance to the Transformer-based methods, ViT~\cite{Dosovitskiy2020An}, ViViT~\cite{Arnab2021ViViTAV} and TubeVit~\cite{Piergiovanni2023RethinkingVV}, due to the effective spatial-temporal modeling of convolutional LSTM units. Among the unimodality-based methods using visual input, our FmFormer-B achieves the best comprehensive performance.

\begin{figure*}
	\begin{center}
		\includegraphics[width=1\linewidth]{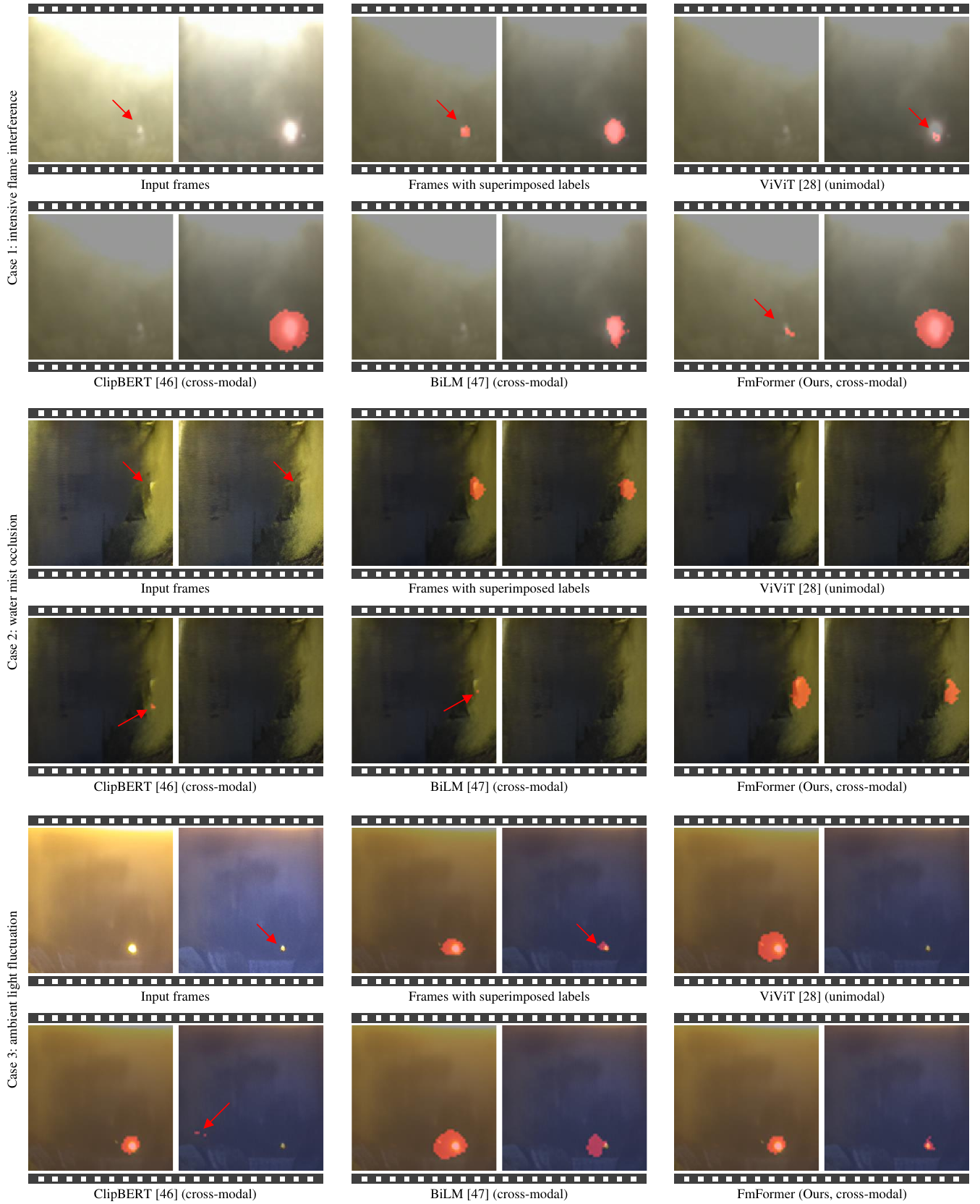}
	\end{center}
	\vspace{-4mm}
	\caption{Visual comparison of the proposed FmFormer-B (cross-modality) with three state-of-the-art Transformer-based methods for pixel-level anomaly detection on three challenging cases. In each case, two close frames in a video are displayed to demonstrate the abnormal dynamics. We superimpose the pixel-level detection result of each method with the corresponding frame for better viewing. In the first case, intense flame light of the furnace (first frame) interferes with most detection methods. The anomaly is not detected until the light interference subsided (second frame). In the second case, visual occlusion from heavy water mist affects the compared methods, resulting in miss detection of the subtle anomaly (second frame). In the third case, fluctuations in ambient light influence the accuracy of anomaly location of the compared methods. In these challenging cases, the proposed FmFormer-B considers current information as a prompt for normal or abnormal conditions, thus demonstrating better robustness under disturbances and temporal consistency under occlusions.}
	\label{fig:comparison}
\end{figure*}

\subsubsection{Cross-modality}
Six state-of-the-art methods are evaluated, including DCNN-SVM~\cite{Zhou2022IdentificationOA}, SSFGGAN~\cite{Bu2023ProcessOP}, Unicoder-VL~\cite{Li2020UnicoderVLAU}, ClipBERT~\cite{Lei2021LessIM}, CAPTURE~\cite{Zhan2021Product1MTW} and BiLM~\cite{yang2022zero}. DCNN-SVM~\cite{Zhou2022IdentificationOA} and SSFGGAN~\cite{Bu2023ProcessOP} achieve multi-modal learning through linear fusion of features from different modalities, which are designed specifically for anomaly detection in fused magnesium smelting processes. Unicoder-VL~\cite{Li2020UnicoderVLAU} encodes vectorized 2D image patches conjointly with 1D sequence (current) into Transformer backbones. ClipBERT~\cite{Lei2021LessIM} and BiLM~\cite{yang2022zero} first employ pre-trained backbones to extract visual features from the video modality, and then embed the visual features together with 1D sequence into Transformers. The different is that ClipBERT~\cite{Lei2021LessIM} squeezes the temporal dimension of visual features via an average-pooling for efficiency. CAPTURE~\cite{Zhan2021Product1MTW} also adopts a pipeline of self-attention and cross-attention for cross-modal learning. Since most baseline methods are designed for classification task, we extend these Transformer-based models to dense prediction task by assembling tokens into image-like features as introduced in Section~\ref{sec:densehead}.

As indicated by the quantitative results in Table~\ref{table:exp1}, the Transformer-based models using cross-modal input (Unicoder-VL~\cite{Li2020UnicoderVLAU}, ClipBERT~\cite{Lei2021LessIM} and BiLM~\cite{yang2022zero}) generally outperform models using single visual modality (ViT~\cite{Dosovitskiy2020An}, ViViT~\cite{Arnab2021ViViTAV} and TubeVit~\cite{Piergiovanni2023RethinkingVV}).
Compared with conventional CNN-based methods using linear fusion (DCNN-SVM~\cite{Zhou2022IdentificationOA} and SSFGGAN~\cite{Bu2023ProcessOP}), Transformer-based methods (Unicoder-VL~\cite{Li2020UnicoderVLAU}, ClipBERT~\cite{Lei2021LessIM} and BiLM~\cite{yang2022zero}) generally demonstrates higher performance, as shown in Table~\ref{table:exp1} since the explicit correlation modeling achieved via Transformer encoder. 
Among methods using cross-modal input, the proposed FmFormer-B achieves superior performances, confirming its effectiveness for both class-level prediction and pixel-level prediction. Moreover, the results indicate that while the video modality provides significant anomaly features, the current modality is crucial for enhancing anomaly detection in the fused magnesium smelting process.

\subsubsection{Dense prediction}
Fig.~\ref{fig:comparison} visualizes the qualitative comparison of dense predictions of the proposed FmFormer-B with three state-of-the-art Transformer-based methods, ViViT~\cite{Arnab2021ViViTAV} (visual), ClipBERT~\cite{Lei2021LessIM} (cross-modal), and BiLM~\cite{yang2022zero} (cross-modal), on three representative cases. In the first case, intense flame light of the furnace has stronger visual saliency than the anomaly region (please see the first frame). The compared methods fail to detect the anomaly under the strong disturbance until the disturbance subsides (second frame). In the second case, heavy water mist occludes the visual features of the anomaly, so the detection method can only rely on the information from the previous frames and three-phase alternating current. In this case, ViViT~\cite{Arnab2021ViViTAV} with only visual input fails to detect the anomaly. In the first frame, ClipBERT~\cite{Lei2021LessIM} and BiLM~\cite{yang2022zero} with cross-modal input detect the anomaly successfully, but predict the wrong location. In the second frame, they also fail to spot the abnormal condition. In the third case, fluctuations of ambient light lead to variations of visual features, which in turn cause the compared methods to misestimate the anomaly regions (see ClipBERT~\cite{Lei2021LessIM} and BiLM~\cite{yang2022zero}) or even produce missed detection (see ViViT~\cite{Arnab2021ViViTAV}).
In comparison, the proposed FmFormer-B leverages current information as a prompt for normal or abnormal conditions, thereby demonstrating robustness under disturbances and temporal consistency under occlusions in complex industrial environment.

\begin{table}[t]
\caption{Performance of the FmFormer-B using cross-modal and unimodal input, where ``LF'' denotes the linear fusion in (\ref{eq:late_fusion}).}
\vspace{-3mm}
\label{table:ablations2}
\begin{center}
\begin{tabular}{l p{1.1cm}<{\centering} p{1.1cm}<{\centering} p{1.1cm}<{\centering} p{1.1cm}<{\centering}}
\toprule
Modality & $Acc \uparrow$ & $F1 \uparrow$ & $FDR \downarrow$  & $MDR \downarrow$\\
\hline
Visual & \underline{0.9731} &  \underline{0.9744} & \textbf{0.0224} &  \underline{0.0308} \\
Current &  0.7754 &  0.7688 & 0.1448 &  0.2954 \\
Cross-modality+LF &  \textbf{0.9748} & \textbf{0.9764} &  \underline{0.0354} &  \textbf{0.0159} \\
\bottomrule
\end{tabular}
\vspace{-3mm}
\end{center}
\end{table}

\begin{table}[t]
\caption{Performance with respect to model capacity (configuration).}
\vspace{-3mm}
\label{table:performance_config}
\begin{center}
\begin{tabular}{l p{1.3cm}<{\centering} p{1.3cm}<{\centering} p{1.3cm}<{\centering} p{1.3cm}<{\centering}}
\toprule
 & $Acc \uparrow$ & Param. \# & FLOPs & FPS $\uparrow$\\
\hline
FmFormer-T & 0.9676 &  0.28M & 110M & 91.5  \\
FmFormer-S &  0.9778 & 0.48M & 174M & 91.2  \\
FmFormer-B &  0.9837 & 1.86M & 610M & 80.1  \\
FmFormer-L &  0.9839 & 3.64M & 1,205M & 48.7 \\
\bottomrule
\vspace{-7mm}
\end{tabular}
\end{center}
\end{table}

\subsection{Ablation Study}
In this section, we empirically analyse the proposed FmFormer by performing the following ablation studies.

\subsubsection{Unimodality vs. cross-modality}\label{Sec:singleVScross}
To investigate the influence of each modality on the performance of anomaly detection, we degrade the proposed FmFormer to a unimodal method by using only one modality as input. Due to the absence of cross-modal input, we replace the MHCA with the MHSA in the Transformer encoder and remove the LF mechanism in (\ref{eq:late_fusion}). Table~\ref{table:ablations2} lists the performance of the proposed FmFormer-B using a single visual or current modality. As expected, the visual model outperforms the current model because the visual features are more stable than the current features.

In this study, the performance of adopting cross-modal input is also verified by simply integrating the two models with unimodal inputs through a LF mechanism in (\ref{eq:late_fusion}). Note that no further training or fine-tuning is performed for the LF mechanism. As shown in Table~\ref{table:ablations2}, despite the poor performance of the current model, it can be integrated with the visual model to improve the comprehensive performance, especially the MDR decreased by nearly half compared to the vision model. This ablation study fully verifies that cross-modal input can effectively enhance anomaly detection.

\subsubsection{Performance with respect to model capacity}\label{sec:performanceVScapacity}
We investigate the performance of our models under different configurations (model capacities), with detailed specifications listed in Table~\ref{table:model_config}. As shown in Table~\ref{table:performance_config}, the prediction accuracy improves from 0.9676 (FmFormer-T) to 0.9839 (FmFormer-L), demonstrating the scalability of FmFormer. Additionally, the inference latency, tested on an NVIDIA TESLA V100, remains satisfactory, highlighting the potential for industrial deployment.

\begin{figure}[t]
	\begin{center}
		\includegraphics[width=0.99\linewidth]{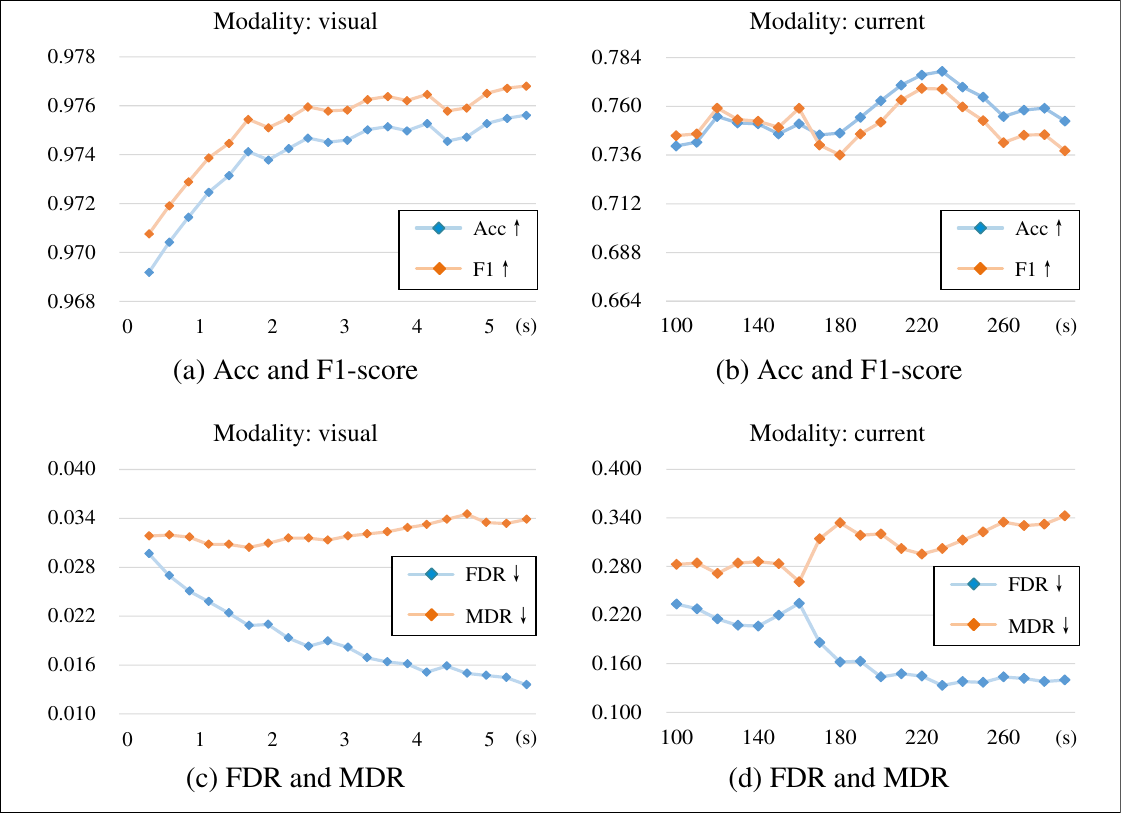}
	\end{center}
 \vspace{-3mm}
	\caption{Performance of the FmFormer-B under different settings of input sequence lengths (horizontal coordinate).}
 \vspace{-1mm}
	\label{fig:sequence_length}
\end{figure}


\begin{table}[t]
\caption{Performance of the FmFormer-B without and with the dilated tokenization mechanism.}
\vspace{-3mm}
\label{table:ablations_token}
\begin{center}
\begin{tabular}{p{1.5cm} p{2.5cm}<{\centering} p{1.5cm}<{\centering} p{1.5cm}<{\centering}}
\toprule
Modality & Dilated tokenization & $Acc \uparrow$ & $mIoU (\%)\uparrow$\\
\hline
\multirow{2}*{Visual} & \ding{55} & 0.9694   & 0.8369\\
& \checkmark &  \textbf{0.9731}  & \textbf{0.8382}\\
\cline{2-4}
\multirow{2}*{Cross-modal} & \ding{55} & 0.9757   & 0.8376\\
& \checkmark &  \textbf{0.9837}  & \textbf{0.8409} \\
\bottomrule
\end{tabular}
\vspace{-3mm}
\end{center}
\end{table}

\subsubsection{Sequence length}
Benefiting from the designed structure, our FmFormer can be adapted to input of variable sequence length without network retraining or fine-tuning\footnote{We use bilinear interpolation for the positional embeddings to accommodate inputs of different sequence lengths.}.
It should be noted that since the input sequences of video and current represent the operating state at the current instant and how that state may change, we only need to ensure the final frame and three-phase current values in the input sequences belong to the same instant, while the sequence lengths of the two modalities do not need to be exactly the same.
The length of the input sequence of each modality influences the network to perceive the dynamic features of normal or abnormal conditions.


\begin{table}[t]
\caption{Performance of the FmFormer-B under different settings of the cross-modal Transformer encoder.}
\vspace{-3mm}
\label{table:ablation_encoder}
\begin{center}
\begin{tabular}{c c cc p{1.1cm}<{\centering} p{1.5cm}<{\centering}}
\toprule
ID & \scriptsize{MHSA} & \scriptsize{MHCA (uni.)} & \scriptsize{MHCA (bi.)} & $Acc \uparrow$ & $mIoU (\%)\uparrow$\\
\hline
0 & \checkmark & \ding{55} & \ding{55} &  0.9748  & \underline{0.8382} \\
1 & \ding{55}  & \checkmark & \ding{55}  &  0.9694   & 0.7562 \\
2 & \ding{55}  & \ding{55}  & \checkmark &  0.9711  & 0.7627 \\
3 & \checkmark & \checkmark & \ding{55} &  \underline{0.9761} & 0.8365\\
4 & \checkmark & \ding{55} & \checkmark & \textbf{0.9837}  & \textbf{0.8409} \\
\bottomrule
\end{tabular}
\end{center}
\vspace{-4mm}
\end{table}

In this experiment, we investigate the performance of our FmFormer-B under different settings of input sequence lengths, as shown in Fig~\ref{fig:sequence_length}.
For the visual modality, the performance of the proposed method continues to increase when the video sequence length is longer than 1.5 seconds. However, the MDR tends to increase when the input sequence is too long. So we chose an input sequence length of 1.5 seconds.
For the current modality, the performance of the proposed method improves as the sequence length increases since it is less influenced by the current noise. 
However, excessively long sequences of current input may drive the self-attention mechanism to tend to focus on the working conditions of other time series instead of the current moment, resulting in the performance degradation. In this experiment, the best performance is achieved when the length of the input current sequence is around 220 seconds.

\subsubsection{Effectiveness of multiscale tokenization}
In this experiment, the effectiveness of the proposed multiscale tokenization module is studied by comparing the results without and with the dilated tokenization mechanism. Note that the multiscale feature blending formulated in (\ref{eq:fusion}) is also removed when the dilated tokenization mechanism is not utilized. As shown in Table~\ref{table:ablations_token}, the models using the proposed multiscale tokenization module achieve overall better performance in both single (visual) modality and cross-modal input settings, which demonstrate the effectiveness of the proposed module.


\subsubsection{Effectiveness of cross-modal Transformer encoder}
We validate the effectiveness of our encoder, which includes the MHSA for internal features encoding within each modality, the MHCA for correlation feature exploration across modalities. The MHCA mechanism includes a unidirectional form, referred to as ``MHCA (uni.)'' for short, and a bidirectional form, denoted as ``MHCA (bi.).'' For configurations using only MHSA or MHCA, we replicate the interaction mechanism to maintain consistent model parameters. Results in Table~\ref{table:ablation_encoder} show that the model with only bidirectional MHCA (ID 2) outperforms the unidirectional (current-to-visual) version (ID 1). However, both forms of solely MHCA perform worse than the model using only MHSA (ID 0), particularly in dense prediction tasks. When combining MHSA with bidirectional MHCA (ID 4), we observe higher accuracy in both classification and dense prediction compared to the unidirectional model (ID 3).

These findings indicate that i) the bidirectional MHCA mechanism effectively enhances class-level and pixel-level prediction tasks; and ii) directly modeling the cross-modal correlation by using cross-attention without self-attention feature encoding degrades the model performance. The self-attention provides sufficient preparation for correlation exploration in cross-attention.

\subsubsection{Effectiveness of multi-head decoder}
We also evaluate various decoder designs, including the classification head (abbreviated as ``cls. head'') with or without the LF mechanism, as well as the dense prediction head (abbreviated as ``dense head'')), as shown in Table~\ref{table:ablation_decoder}. 
Using only the single dense prediction head (ID 0), our FmFormer-B suffers a performance degradation in mIoU. Additionally, the single classification head (ID 1) performs worse compared to our full model with the multi-head decoder (ID3). These performance losses are essentially due to training with a single loss term ($\mathcal{L}^{pix}$ or $\mathcal{L}^{cls}$). Furthermore, when comparing the model without LF mechanism (ID 2) to the version with it (ID 3), the latter demonstrates significantly better performance.
Since the LF mechanism does not interact with the dense prediction head, the mIoU values are the same with or without it.
These results highlight the effectiveness of the proposed multi-head decoder.

\begin{table}[t]
\caption{Performance of the FmFormer-B under different settings of the multi-head decoder.}
\vspace{-6mm}
\label{table:ablation_decoder}
\begin{center}
\begin{tabular}{c c c p{1.1cm}<{\centering} p{1.1cm}<{\centering} p{1.5cm}<{\centering}}
\toprule
ID & Cls. head & Dense head & LF & $Acc \uparrow$ & $mIoU (\%)\uparrow$\\
\hline
0 & \ding{55}  & \checkmark & \ding{55}  &  -   & 0.8276 \\
1 & \checkmark & \ding{55} & \checkmark &  \underline{0.9795}  & - \\
2 & \checkmark & \checkmark & \ding{55} &  0.9757 & \textbf{0.8409}\\
3 & \checkmark & \checkmark & \checkmark & \textbf{0.9837}  & \textbf{0.8409} \\
\bottomrule
\end{tabular}
\end{center}
\vspace{-4mm}
\end{table}

\section{Conclusion and Discussion}\label{sec:conclusion}
In this paper, we introduce a novel Transformer, dubbed FmFormer, through the lens of cross-modal learning to enhance anomaly detection in fused magnesium smelting process. In the proposed FmFormer, a multiscale tokenization module is developed to handle the problem of large dimensionality gap between modalities. A cross-modal Transformer encoder is then employed to alternatively explore the internal features of each modality and the correlation features across modalities. 
Through a multi-head decoder, our FmFormer is able to preform class-level anomaly prediction and pixel-level anomaly region detection. 
Interestingly, despite the poor detection accuracy when using a single current modality, the comprehensive performance of the model can still be improved through a simple linear fusion.
Furthermore, by taking advantage of the cross-modal learning, the proposed method achieves an accurate anomaly detection under extreme interferences such as current fluctuation and visual occlusion from heavy water mist. To demonstrate the effectiveness of the FmFormer, we present the first cross-modal benchmark for anomaly detection of fused magnesium smelting processes, which possesses synchronously acquired video-current data and pixel-level labels. We hope it will be helpful to the research community of cross-modal learning in complex industrial scenarios.

\textbf{Limitations and future works.} 
Our FmFormer is optimized for temporally aligned inputs (i.e., synchronized data), but achieving strict alignment among multiple modalities in real-world industrial applications can be challenging. This may be addressed by leveraging the global perception capabilities of attention mechanisms along the temporal dimension, paired with a learning paradigm for unaligned modalities. Additionally, network communication instability may lead to missing data points or even inaccessible modalities. Implementing a masked cross-modal modeling approach with adaptive modality switching could help, along with joint training that combines image-current, video-current, and unimodal inputs. Currently, our implementation relies solely on current-to-video tokens from dilated and standard tokenization for dense prediction. To further enhance performance, one could also incorporate video-to-current tokens and class tokens.

\begin{IEEEbiography}[{\includegraphics[width=1in]{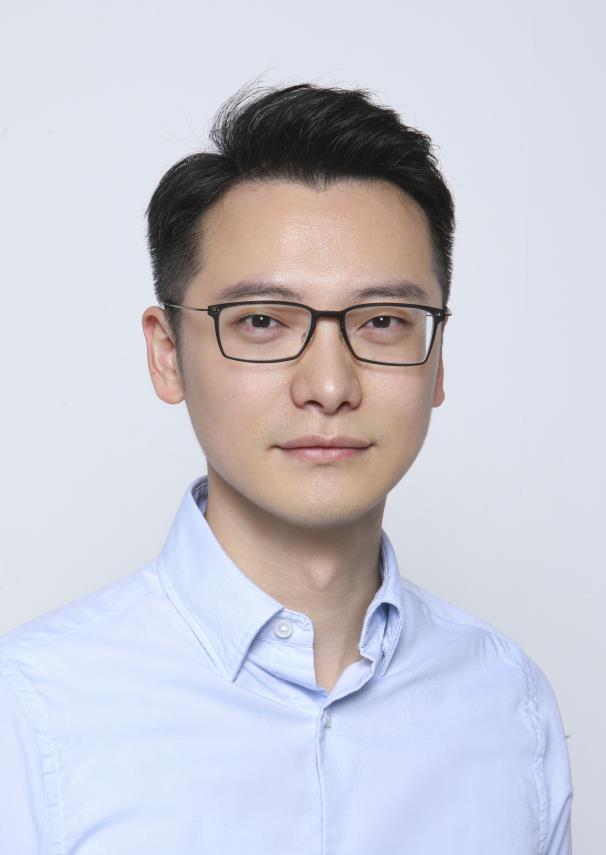}}]{Gaochang Wu}
received the BE and MS degrees in mechanical engineering in Northeastern University, Shenyang, China, in 2013 and 2015, respectively, and Ph.D. degree in control theory and control engineering in Northeastern University, Shenyang, China in 2020. He is currently an associate professor in the State Key Laboratory of Synthetical Automation for Process Industries, Northeastern University. He was selected for the 2022-2024 Youth Talent Support Program of the Chinese Association of Automation. His current research interests include multimodal perception and recognition, light field imaging and processing, and computer vision in industrial scenarios.
\vspace{-5mm}
\end{IEEEbiography}

\begin{IEEEbiography}[{\includegraphics[width=1in]{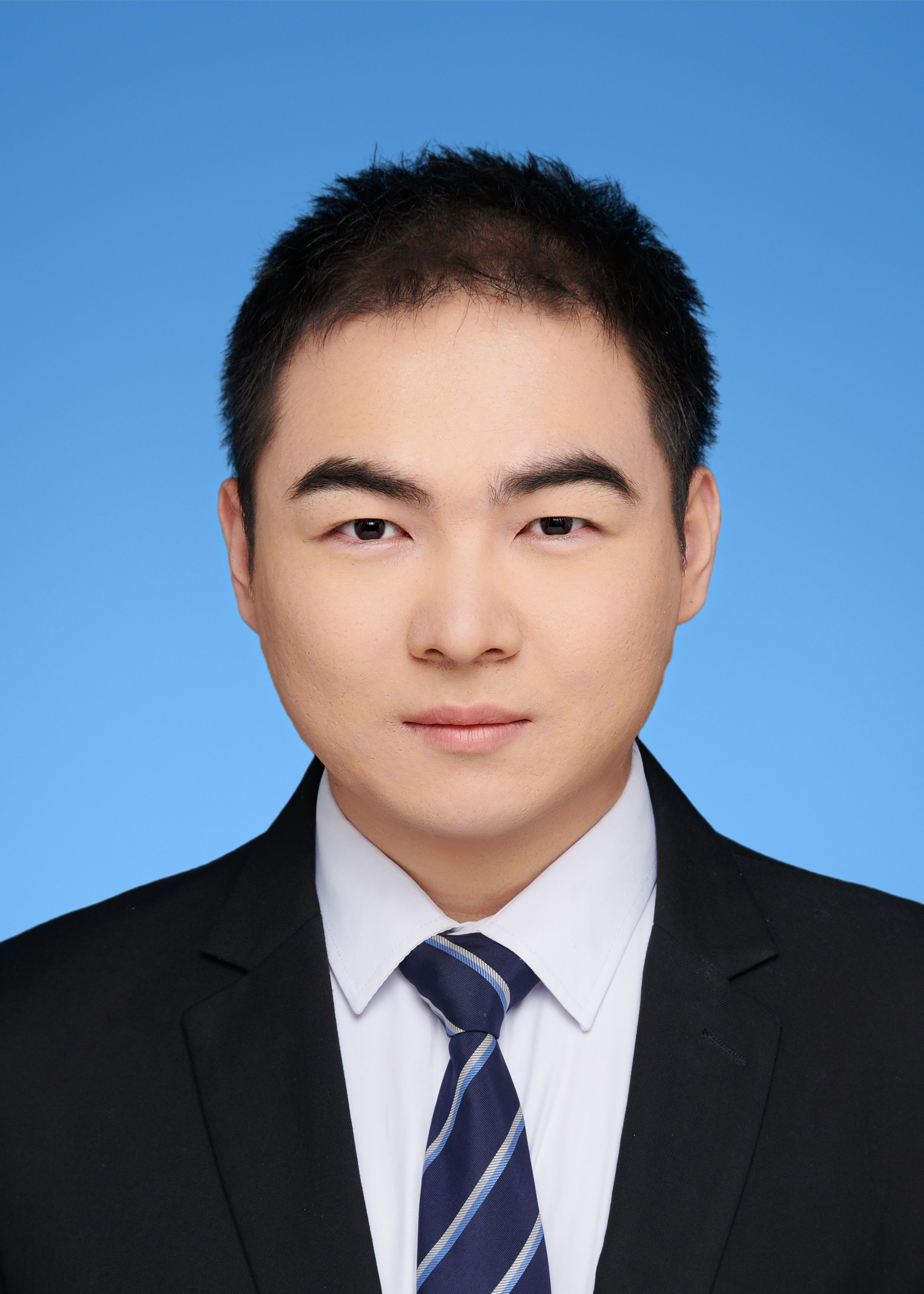}}]{Yapeng Zhang}
received the BE degree in automation from Northeastern University, Shenyang, China, in 2022. He is currently pursuing a MS degree in the State Key Laboratory of Synthetical Automation for Process Industries. His current research focuses on multimodal learning, anomaly detection and computer vision.
\vspace{-5mm}
\end{IEEEbiography}

\begin{IEEEbiography}[{\includegraphics[width=1in]{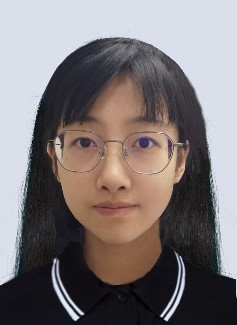}}]{Lan Deng}
received the MS degree in automation from Northeastern University, Shenyang, China, in 2021. She is currently pursuing the Ph.D. degree in the State Key Laboratory of Synthetical Automation for Process Industries. Her main research interest is deep learning, anomaly detection with applications to process industries, and industrial metaverse.
\vspace{-5mm}
\end{IEEEbiography}

\begin{IEEEbiography}[{\includegraphics[width=1in]{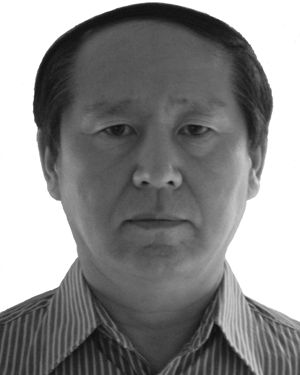}}]{Jingxin Zhang}
received the ME and PhD degrees in electrical engineering from Northeastern University, Shenyang, China. Since 1989, he has held research and academic positions with Northeastern University, China, the University of Florence, Italy, the University of Melbourne, the University of South Australia, Deakin University and Monash University, Australia. He is currently an associate professor of electrical engineering at the Swinburne University of Technology, and an adjunct associate professor of electrical and computer systems engineering at Monash University, Melbourne, Australia. His research interests include signals and systems and their applications to biomedical and industrial systems. He is the recipient of the 1989 Fok Ying Tong Educational Foundation (Hong Kong) for the Outstanding Young Faculty Members in China, and 1992 China National Education Committee Award for the Advancement of Science and Technology.
\vspace{-5mm}
\end{IEEEbiography}

\begin{IEEEbiography}[{\includegraphics[width=1in]{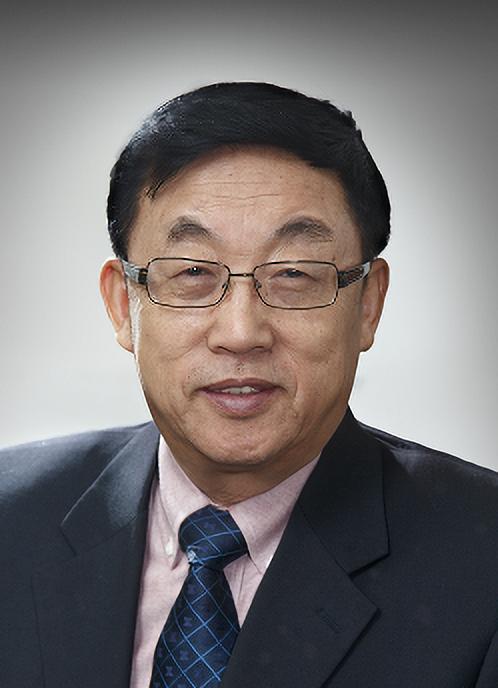}}]{Tianyou Chai}
received the Ph.D. degree in control theory and engineering from Northeastern University, Shenyang, China, in 1985. He has been with the Research Center of Automation, Northeastern University, Shenyang, China, since 1985, where he became a Professor in 1988 and a Chair Professor in 2004. His current research interests include adaptive control, intelligent decoupling control, integrated plant control and systems, and the development of control technologies with applications to various industrial processes. Prof. Chai is a member of the Chinese Academy of Engineering, an academician of International Eurasian Academy of Sciences, IEEE Fellow and IFAC Fellow. He is a distinguished visiting fellow of The Royal Academy of Engineering (UK) and an Invitation Fellow of Japan Society for the Promotion of Science (JSPS).
\end{IEEEbiography}

\vspace{11pt}

\vfill


\begin{thebibliography}{10}
\bibitem{Wu2018NonlinearCT}
Z.~Wu, T.~Liu, Z.-P. Jiang, T.~you Chai, and L.~Zhang, ``Nonlinear control tools for fused magnesium furnaces: Design and implementation,'' \emph{IEEE Transactions on Industrial Electronics}, vol.~65, pp. 7248--7257, 2018.

\bibitem{Chai2017ACB}
T.~you Chai, Z.~Wu, and H.~Wang, ``A cps based optimal operational control system for fused magnesium furnace,'' \emph{IFAC-PapersOnLine}, vol.~50, pp. 14\,992--14\,999, 2017.

\bibitem{Wu2015DataDrivenAC}
Z.~Wu, Y.~Wu, T.~you Chai, and J.~Sun, ``Data-driven abnormal condition identification and self-healing control system for fused magnesium furnace,'' \emph{IEEE Transactions on Industrial Electronics}, vol.~62, pp. 1703--1715, 2015.

\bibitem{Zhang2011FaultDO}
Y.-W. Zhang and C.~Ma, ``Fault diagnosis of nonlinear processes using multiscale kpca and multiscale kpls,'' \emph{Chemical Engineering Science}, vol.~66, pp. 64--72, 2011.

\bibitem{Wang2023DensityBasedSP}
Z.~Wang and Y.~Fan, ``Density-based structure preserving projections process monitoring model for fused magnesia smelting process,'' \emph{IEEE Transactions on Industrial Informatics}, vol.~19, pp. 9654--9666, 2023.

\bibitem{LeCun2015DeepL}
Y.~LeCun, Y.~Bengio, and G.~E. Hinton, ``Deep learning,'' \emph{Nature}, vol. 521, pp. 436--444, 2015.

\bibitem{Silver2017MasteringTG}
D.~Silver, J.~Schrittwieser, K.~Simonyan, I.~Antonoglou, A.~Huang, A.~Guez, T.~Hubert, L.~baker, M.~Lai, A.~Bolton, Y.~Chen, T.~P. Lillicrap, F.~Hui, L.~Sifre, G.~van~den Driessche, T.~Graepel, and D.~Hassabis, ``Mastering the game of go without human knowledge,'' \emph{Nature}, vol. 550, pp. 354--359, 2017.

\bibitem{Zhou2022IdentificationOA}
P.~Zhou, B.~Gao, S.~Wang, and T.~you Chai, ``Identification of abnormal conditions for fused magnesium melting process based on deep learning and multisource information fusion,'' \emph{IEEE Transactions on Industrial Electronics}, vol.~69, pp. 3017--3026, 2022.

\bibitem{Lu2022SemiSupervisedCM}
S.~Lu and Y.~Wen, ``Semi-supervised condition monitoring and visualization of fused magnesium furnace,'' \emph{IEEE Transactions on Automation Science and Engineering}, vol.~19, pp. 3471--3482, 2022.

\bibitem{10587314}
Q.~Yu, K.~Zhu, Y.~Cao, F.~Xia, and Y.~Kang, ``Tf2: Few-shot text-free training-free defect image generation for industrial anomaly inspection,'' \emph{IEEE Transactions on Circuits and Systems for Video Technology}, pp. 1--1, 2024.

\bibitem{wu2019abnormal}
G.~Wu, Q.~Liu, T.~Chai, and S.~Qin, ``Abnormal condition diagnosis through deep learning of image sequences for fused magnesium furnaces,'' \emph{Acta Automatica Sinica}, vol.~45, no.~8, pp. 1475--1485, 2019.

\bibitem{Liu2023DisturbanceRA}
Q.~Liu, Y.~Zhang, G.~Wu, and Z.~Fan, ``Disturbance robust abnormality diagnosis of fused magnesium furnaces using deep neural networks,'' \emph{IEEE Transactions on Artificial Intelligence}, vol.~4, pp. 669--678, 2023.

\bibitem{Hochreiter1997LongSM}
S.~Hochreiter and J.~Schmidhuber, ``Long short-term memory,'' \emph{Neural Computation}, vol.~9, pp. 1735--1780, 1997.

\bibitem{Vaswani2017AttentionIA}
A.~Vaswani, N.~M. Shazeer, N.~Parmar, J.~Uszkoreit, L.~Jones, A.~N. Gomez, L.~Kaiser, and I.~Polosukhin, ``Attention is all you need,'' in \emph{Neural Information Processing Systems}, 2017.

\bibitem{OpenAI2023GPT4TR}
OpenAI, ``Gpt-4 technical report,'' \emph{ArXiv}, vol. abs/2303.08774, 2023.

\bibitem{Dosovitskiy2020An}
A.~Dosovitskiy, L.~Beyer, A.~Kolesnikov, D.~Weissenborn, X.~Zhai, T.~Unterthiner, M.~Dehghani, M.~Minderer, G.~Heigold, S.~Gelly, J.~Uszkoreit, and N.~Houlsby, ``An image is worth 16x16 words: Transformers for image recognition at scale,'' in \emph{International Conference on Learning Representations}, 2020.

\bibitem{Ni2022ExpandingLP}
B.~Ni, H.~Peng, M.~Chen, S.~Zhang, G.~Meng, J.~Fu, S.~Xiang, and H.~Ling, ``Expanding language-image pretrained models for general video recognition,'' in \emph{European Conference on Computer Vision}, 2022.

\bibitem{Driess2023PaLMEAE}
D.~Driess, F.~Xia, M.~S.~M. Sajjadi, C.~Lynch, A.~Chowdhery, B.~Ichter, A.~Wahid, J.~Tompson, Q.~H. Vuong, T.~Yu, W.~Huang, Y.~Chebotar, P.~Sermanet, D.~Duckworth, S.~Levine, V.~Vanhoucke, K.~Hausman, M.~Toussaint, K.~Greff, A.~Zeng, I.~Mordatch, and P.~R. Florence, ``Palm-e: An embodied multimodal language model,'' in \emph{International Conference on Machine Learning}, 2023.

\bibitem{Wang2023ActionCLIP}
M.~Wang, J.~Xing, J.~Mei, Y.~Liu, and Y.~Jiang, ``Actionclip: Adapting language-image pretrained models for video action recognition,'' \emph{IEEE Transactions on Neural Networks and Learning Systems}, pp. 1--13, 2023.

\bibitem{Chandola2009AnomalyDA}
V.~Chandola, A.~Banerjee, and V.~Kumar, ``Anomaly detection: A survey,'' \emph{ACM Comput. Surv.}, vol.~41, pp. 15:1--15:58, 2009.

\bibitem{Yin2022AnomalyDB}
C.~Yin, S.~Zhang, J.~Wang, and N.~N. Xiong, ``Anomaly detection based on convolutional recurrent autoencoder for iot time series,'' \emph{IEEE Transactions on Systems, Man, and Cybernetics: Systems}, vol.~52, pp. 112--122, 2022.

\bibitem{Barz2018DetectingRO}
B.~Barz, E.~Rodner, Y.~G. Garcia, and J.~Denzler, ``Detecting regions of maximal divergence for spatio-temporal anomaly detection,'' \emph{IEEE Transactions on Pattern Analysis and Machine Intelligence}, vol.~41, pp. 1088--1101, 2018.

\bibitem{Liu2023Time}
S.~Liu, B.~Zhou, Q.~Ding, B.~Hooi, Z.~Zhang, H.~Shen, and X.~Cheng, ``Time series anomaly detection with adversarial reconstruction networks,'' \emph{IEEE Transactions on Knowledge and Data Engineering}, vol.~35, no.~4, pp. 4293--4306, 2023.

\bibitem{Zaheer2022GenerativeCL}
M.~Zaheer, A.~Mahmood, M.~H. Khan, M.~Segu, F.~Yu, and S.-I. Lee, ``Generative cooperative learning for unsupervised video anomaly detection,'' \emph{IEEE/CVF Conference on Computer Vision and Pattern Recognition}, pp. 14\,724--14\,734, 2022.

\bibitem{Xu2022Anomaly}
J.~Xu, H.~Wu, J.~Wang, and M.~Long, ``Anomaly transformer: Time series anomaly detection with association discrepancy,'' in \emph{International Conference on Learning Representations}, 2022.

\bibitem{Chen2023BidirectionalSA}
C.~Chen, Y.~Liu, L.~Chen, and C.~Zhang, ``Bidirectional spatial-temporal adaptive transformer for urban traffic flow forecasting,'' \emph{IEEE Transactions on Neural Networks and Learning Systems}, vol.~34, pp. 6913--6925, 2023.

\bibitem{Li2022SelfTrainingML}
S.~Li, F.~Liu, and L.~Jiao, ``Self-training multi-sequence learning with transformer for weakly supervised video anomaly detection,'' in \emph{AAAI Conference on Artificial Intelligence}, 2022.

\bibitem{Arnab2021ViViTAV}
A.~Arnab, M.~Dehghani, G.~Heigold, C.~Sun, M.~Lucic, and C.~Schmid, ``Vivit: A video vision transformer,'' \emph{2021 IEEE/CVF International Conference on Computer Vision}, pp. 6816--6826, 2021.

\bibitem{Piergiovanni2023RethinkingVV}
A.~J. Piergiovanni, W.~Kuo, and A.~Angelova, ``Rethinking video vits: Sparse video tubes for joint image and video learning,'' \emph{2023 IEEE/CVF Conference on Computer Vision and Pattern Recognition}, pp. 2214--2224, 2023.

\bibitem{Yu2015Dilatconv}
F.~Yu and V.~Koltun, ``Multi-scale context aggregation by dilated convolutions.'' in \emph{ICLR}, 2015.

\bibitem{Zhao2018CrossScaleRL}
M.~Zhao, G.~Wu, Y.~Li, X.~Hao, L.~Fang, and Y.~Liu, ``Cross-scale reference-based light field super-resolution,'' \emph{IEEE Transactions on Computational Imaging}, vol.~4, pp. 406--418, 2018.

\bibitem{Xu2020U2FusionAU}
H.~Xu, J.~Ma, J.~Jiang, X.~Guo, and H.~Ling, ``U2fusion: A unified unsupervised image fusion network,'' \emph{IEEE Transactions on Pattern Analysis and Machine Intelligence}, vol.~44, pp. 502--518, 2020.

\bibitem{Zhou2021CrossMPICS}
Y.~Zhou, G.~Wu, Y.~Fu, K.~Li, and Y.~Liu, ``Cross-mpi: Cross-scale stereo for image super-resolution using multiplane images,'' \emph{IEEE/CVF Conference on Computer Vision and Pattern Recognition}, pp. 14\,837--14\,846, 2021.

\bibitem{Shao2021LocalTransAM}
R.~Shao, G.~Wu, Y.~Zhou, Y.~Fu, and Y.~Liu, ``Localtrans: A multiscale local transformer network for cross-resolution homography estimation,'' \emph{IEEE/CVF International Conference on Computer Vision}, pp. 14\,870--14\,879, 2021.

\bibitem{10163247}
S.~Park, A.~G. Vien, and C.~Lee, ``Cross-modal transformers for infrared and visible image fusion,'' \emph{IEEE Transactions on Circuits and Systems for Video Technology}, vol.~34, no.~2, pp. 770--785, 2024.

\bibitem{Benyounes2017MUTANMT}
H.~Ben-younes, R.~Cad{\`e}ne, M.~Cord, and N.~Thome, ``Mutan: Multimodal tucker fusion for visual question answering,'' \emph{2017 IEEE International Conference on Computer Vision}, pp. 2631--2639, 2017.

\bibitem{Ju2021PromptingVM}
C.~Ju, T.~Han, K.~Zheng, Y.~Zhang, and W.~Xie, ``Prompting visual-language models for efficient video understanding,'' in \emph{European Conference on Computer Vision}, 2022, pp. 105--124.

\bibitem{Wu2022BidirectionalCK}
W.~Wu, X.~Wang, H.~Luo, J.~Wang, Y.~Yang, and W.~Ouyang, ``Bidirectional cross-modal knowledge exploration for video recognition with pre-trained vision-language models,'' \emph{2023 IEEE/CVF Conference on Computer Vision and Pattern Recognition}, pp. 6620--6630, 2022.

\bibitem{10057259}
R.-C. Tu, J.~Jiang, Q.~Lin, C.~Cai, S.~Tian, H.~Wang, and W.~Liu, ``Unsupervised cross-modal hashing with modality-interaction,'' \emph{IEEE Transactions on Circuits and Systems for Video Technology}, vol.~33, no.~9, pp. 5296--5308, 2023.

\bibitem{10445315}
Z.~Li, L.~Zhang, K.~Zhang, Y.~Zhang, and Z.~Mao, ``Improving image-text matching with bidirectional consistency of cross-modal alignment,'' \emph{IEEE Transactions on Circuits and Systems for Video Technology}, vol.~34, no.~7, pp. 6590--6607, 2024.

\bibitem{10345597}
Y.~Wang, Y.-W. Zhan, Z.-D. Chen, X.~Luo, and X.-S. Xu, ``Multiple information embedded hashing for large-scale cross-modal retrieval,'' \emph{IEEE Transactions on Circuits and Systems for Video Technology}, vol.~34, no.~6, pp. 5118--5131, 2024.

\bibitem{Yang2021LearningVQ}
R.~Yang, M.~Zhang, N.~Hansen, H.~Xu, and X.~Wang, ``Learning vision-guided quadrupedal locomotion end-to-end with cross-modal transformers,'' \emph{ArXiv}, vol. abs/2107.03996, 2021.

\bibitem{Brown2020LanguageMA}
T.~B. Brown, B.~Mann, N.~Ryder, M.~Subbiah, J.~Kaplan, P.~Dhariwal, A.~Neelakantan, P.~Shyam, G.~Sastry, A.~Askell, S.~Agarwal, A.~Herbert-Voss, G.~Krueger, T.~J. Henighan, R.~Child, A.~Ramesh, D.~M. Ziegler, J.~Wu, C.~Winter, C.~Hesse, M.~Chen, E.~Sigler, M.~Litwin, S.~Gray, B.~Chess, J.~Clark, C.~Berner, S.~McCandlish, A.~Radford, I.~Sutskever, and D.~Amodei, ``Language models are few-shot learners,'' in \emph{Conference on Neural Information Processing Systems}, 2020, pp. 1877--1901.

\bibitem{Bu2023ProcessOP}
K.~Bu, Y.~Liu, and F.~Wang, ``Process operation performance assessment based on semi-supervised fine-grained generative adversarial network for efmf,'' \emph{IEEE Transactions on Instrumentation and Measurement}, vol.~72, pp. 1--9, 2023.

\bibitem{Xu2023MultimodalLW}
P.~Xu, X.~Zhu, and D.~A. Clifton, ``Multimodal learning with transformers: A survey,'' \emph{IEEE Transactions on Pattern Analysis and Machine Intelligence}, vol.~45, pp. 12\,113--12\,132, 2023.

\bibitem{Lei2021LessIM}
J.~Lei, L.~Li, L.~Zhou, Z.~Gan, T.~L. Berg, M.~Bansal, and J.~Liu, ``Less is more: Clipbert for video-and-language learning via sparse sampling,'' \emph{2021 IEEE/CVF Conference on Computer Vision and Pattern Recognition}, pp. 7327--7337, 2021.

\bibitem{yang2022zero}
A.~Yang, A.~Miech, J.~Sivic, I.~Laptev, and C.~Schmid, ``Zero-shot video question answering via frozen bidirectional language models,'' \emph{Advances in Neural Information Processing Systems}, vol.~35, pp. 124--141, 2022.

\bibitem{Radford2021LearningTV}
A.~Radford, J.~W. Kim, C.~Hallacy, A.~Ramesh, G.~Goh, S.~Agarwal, G.~Sastry, A.~Askell, P.~Mishkin, J.~Clark, G.~Krueger, and I.~Sutskever, ``Learning transferable visual models from natural language supervision,'' in \emph{International Conference on Machine Learning}, 2021.

\bibitem{Wu2022RevisitingCT}
W.~Wu, Z.~Sun, and W.~Ouyang, ``Revisiting classifier: Transferring vision-language models for video recognition,'' in \emph{AAAI Conference on Artificial Intelligence}, 2022.

\bibitem{Wu2023TransferringVM}
W.~Wu, Z.~Sun, Y.~Song, J.~Wang, and W.~Ouyang, ``Transferring vision-language models for visual recognition: A classifier perspective,'' \emph{International Journal of Computer Vision}, 2023.

\bibitem{Neimark2021VideoTN}
D.~Neimark, O.~Bar, M.~Zohar, and D.~Asselmann, ``Video transformer network,'' \emph{IEEE/CVF International Conference on Computer Vision Workshops}, pp. 3156--3165, 2021.

\bibitem{Ranftl2021VisionTF}
R.~Ranftl, A.~Bochkovskiy, and V.~Koltun, ``Vision transformers for dense prediction,'' \emph{2021 IEEE/CVF International Conference on Computer Vision}, pp. 12\,159--12\,168, 2021.

\bibitem{Ba2016LayerN}
J.~Ba, J.~R. Kiros, and G.~E. Hinton, ``Layer normalization,'' \emph{ArXiv}, vol. abs/1607.06450, 2016.

\bibitem{Li2021GroundedLP}
L.~H. Li, P.~Zhang, H.~Zhang, J.~Yang, C.~Li, Y.~Zhong, L.~Wang, L.~Yuan, L.~Zhang, J.-N. Hwang, K.-W. Chang, and J.~Gao, ``Grounded language-image pre-training,'' \emph{2022 IEEE/CVF Conference on Computer Vision and Pattern Recognition}, pp. 10\,955--10\,965, 2021.

\bibitem{WMF2014}
Q.~Zhang, L.~Xu, and J.~Jia, ``100+ times faster weighted median filter (wmf),'' \emph{IEEE Conference on Computer Vision and Pattern Recognition}, pp. 2830--2837, 2014.

\bibitem{Paszke2019PyTorchAI}
A.~Paszke, S.~Gross, F.~Massa, A.~Lerer, J.~Bradbury, G.~Chanan, T.~Killeen, Z.~Lin, N.~Gimelshein, L.~Antiga, A.~Desmaison, A.~K{\"o}pf, E.~Yang, Z.~DeVito, M.~Raison, A.~Tejani, S.~Chilamkurthy, B.~Steiner, L.~Fang, J.~Bai, and S.~Chintala, ``Pytorch: An imperative style, high-performance deep learning library,'' in \emph{Neural Information Processing Systems}, 2019.

\bibitem{Loshchilov2017ADAMW}
I.~Loshchilov and F.~Hutter, ``Decoupled weight decay regularization,'' in \emph{International Conference on Learning Representations}, 2017.

\bibitem{Sokolova2009ASA}
M.~Sokolova and G.~Lapalme, ``A systematic analysis of performance measures for classification tasks,'' \emph{Inf. Process. Manag.}, vol.~45, pp. 427--437, 2009.

\bibitem{Everingham2010ThePV}
M.~Everingham, L.~V. Gool, C.~K.~I. Williams, J.~M. Winn, and A.~Zisserman, ``The pascal visual object classes (voc) challenge,'' \emph{International Journal of Computer Vision}, vol.~88, pp. 303--338, 2010.

\bibitem{Zhou2020InformerBE}
H.~Zhou, S.~Zhang, J.~Peng, S.~Zhang, J.~Li, H.~Xiong, and W.~Zhang, ``Informer: Beyond efficient transformer for long sequence time-series forecasting,'' in \emph{AAAI Conference on Artificial Intelligence}, 2020.

\bibitem{Wu2022FlowformerLT}
H.~Wu, J.~Wu, J.~Xu, J.~Wang, and M.~Long, ``Flowformer: Linearizing transformers with conservation flows,'' in \emph{International Conference on Machine Learning}, 2022.

\bibitem{Dao2022FlashAttentionFA}
T.~Dao, D.~Y. Fu, S.~Ermon, A.~Rudra, and C.~R'e, ``Flashattention: Fast and memory-efficient exact attention with io-awareness,'' in \emph{Neural Information Processing Systems}, 2022.

\bibitem{liu2023itransformer}
Y.~Liu, T.~Hu, H.~Zhang, H.~Wu, S.~Wang, L.~Ma, and M.~Long, ``itransformer: Inverted transformers are effective for time series forecasting,'' \emph{arXiv preprint arXiv:2310.06625}, 2023.

\bibitem{Li2020UnicoderVLAU}
G.~Li, N.~Duan, Y.~Fang, D.~Jiang, and M.~Zhou, ``Unicoder-vl: A universal encoder for vision and language by cross-modal pre-training,'' in \emph{AAAI Conference on Artificial Intelligence}, 2020.

\bibitem{Zhan2021Product1MTW}
X.~Zhan, Y.~Wu, X.~Dong, Y.~Wei, M.~Lu, Y.~Zhang, H.~Xu, and X.~Liang, ``Product1m: Towards weakly supervised instance-level product retrieval via cross-modal pretraining,'' \emph{2021 IEEE/CVF International Conference on Computer Vision}, pp. 11\,762--11\,771, 2021.

\end{thebibliography}
\end{document}